\documentclass{article}
\usepackage{ijcai21}

\usepackage{times}

\usepackage{soul}
\usepackage{url}
\usepackage[hidelinks]{hyperref}
\usepackage[utf8]{inputenc}
\usepackage[small]{caption}
\usepackage{graphicx}
\usepackage{amsmath}
\usepackage{booktabs}
\usepackage{natbib}
\usepackage{scalerel}
\usepackage{amssymb}
\usepackage{graphics}
\usepackage{subcaption}
\usepackage{dsfont}
\usepackage{algorithm}
\usepackage{algorithmic}
\usepackage{upgreek}
\usepackage{color}
\usepackage{multirow}
\usepackage{listings}

\usepackage{amssymb}
\usepackage{amsmath,amsfonts}
\usepackage{amsopn,bm,mathtools}

\usepackage{nicefrac}       % compact symbols for 1/2, etc.
\newcommand{\ProbOpr}[1]{\mathbb{#1}}
\newcommand{\expect}[2]{%
\ifthenelse{\equal{#2}{}}{\ProbOpr{E}_{#1}}
{\ifthenelse{\equal{#1}{}}{\ProbOpr{E}\left[#2\right]}{\ProbOpr{E}_{#1}\left[#2\right]}}} % Expectation: syntax: E{1}{2} = E_1[2], E{}{2}=E[2], E{1}{} = E_1
\newcommand{\var}[2]{%
\ifthenelse{\equal{#2}{}}{\ProbOpr{VAR}_{#1}}
{\ifthenelse{\equal{#1}{}}{\ProbOpr{VAR}\left[#2\right]}{\ProbOpr{VAR}_{#1}\left[#2\right]}}} % Expectation: syntax: V{1}{2} = V_1[2], V{}{2}=V[2], V{1}{} = V_1
\usepackage{xspace}
\makeatletter
\DeclareRobustCommand\onedot{\futurelet\@let@token\@onedot}
\def\@onedot{\ifx\@let@token.\else.\null\fi\xspace}

\makeatother

\usepackage{mathtools}

\usepackage{listings}

\graphicspath{{../figs/}{../}}

\definecolor{dkgreen}{rgb}{0,0.6,0}
\definecolor{gray}{rgb}{0.5,0.5,0.5}
\definecolor{mauve}{rgb}{0.58,0,0.82}

\title{Coordinated Exploration via Intrinsic Rewards for Multi-Agent Reinforcement Learning}

\author{
Shariq Iqbal$^1$ \and
Fei Sha $^{1,2}$
\affiliations
$^1$ University of Southern California \\
$^2$ Google AI
}

\begin{document}

\maketitle

\begin{abstract}
Solving tasks with sparse rewards is one of the most important challenges in reinforcement learning.
In the single-agent setting, this challenge is addressed by introducing intrinsic rewards that motivate agents to explore unseen regions of their state spaces; however, applying these techniques naively to the multi-agent setting results in agents exploring independently, without any coordination among themselves.
Exploration in cooperative multi-agent settings can be accelerated and improved if agents coordinate their exploration.
In this paper we introduce a framework for designing intrinsic rewards which consider what other agents have explored such that the agents can coordinate.
Then, we develop an approach for learning how to dynamically select between several exploration modalities to maximize extrinsic rewards.
Concretely, we formulate the approach as a hierarchical policy where a high-level controller selects among sets of policies trained on diverse intrinsic rewards and the low-level controllers learn the action policies of all agents under these specific rewards.
We demonstrate the effectiveness of the proposed approach in cooperative domains with sparse rewards where state-of-the-art methods fail and challenging multi-stage tasks that necessitate changing modes of coordination.
\end{abstract}

\section{Introduction}
\label{intro}
%!TEX root = ms.tex
Solving tasks with sparse rewards is a fundamental challenge of reinforcement learning.
This challenge is most commonly addressed by learning with intrinsic rewards that encourage exploration of the state space~\citep{pmlr-v70-pathak17a,houthooft2016vime,burda2018exploration,ostrovski2017count,tang2017exploration}.
In the cooperative multi-agent setting, the sparse reward challenge is exacerbated by the need for agents to \emph{coordinate} their exploration.
In many cases, the non-coordinated approach -- agents exploring independently -- is not efficient.
For example, consider a search-and-rescue task where multiple agents need to collectively find all missing persons spread throughout their environment and bring them to a common recovery location.
During the search phase, it would be inefficient for the agents to explore the same areas redundantly.
Instead, it would be much more sensible for agents to ``divide-and-conquer'' or avoid redundant exploration.
Thus, an ideal intrinsic reward for this phase would encourage such behavior; however, the same behavior would not be ideal during the recovery phase where agents must converge at a common location.
Cooperative multi-agent reinforcement learning can benefit from coordinating exploration across agents; however, the type of coordination should be \emph{adaptive} to the task at hand.

In this work, we introduce a framework for designing multi-agent intrinsic rewards that coordinate with respect to explored regions, then we present a method for learning both low-level policies trained on different intrinsic rewards and a meta-policy for selecting the policies which maximize extrinsic rewards on a given task\footnote{Code available at: \href{https://github.com/shariqiqbal2810/Multi-Explore}{https://github.com/shariqiqbal2810/Multi-Explore}}.
Importantly, we learn the policies simultaneously using a shared replay buffer with off-policy methods, drastically improving sample efficiency.
This shared replay buffer enables us to use \emph{all} data to train \emph{all} policies, rather than needing to collect data with the specific policies we want to update.
Moreover, the meta-policy is learned in conjunction with those low-level policies, effectively exploring over the space of coordinated low-level exploration types.
We show empirically, in both a GridWorld domain as well as in the more complex 3D ViZDoom~\citep{Kempka2016ViZDoom} setting: 1) intrinsic reward functions which coordinate across agents are more effective than independent intrinsically motivated exploration, 2) our approach is able to match or exceed the performance of the best coordinated intrinsic reward function (which differs across tasks) while using no more samples, and 3) on challenging multi-stage tasks requiring varying modes of cooperation, our adaptive approach outperforms all individual reward functions and continues to learn while the other approaches stagnate due to their lack of coordination and/or adaptability.

\section{Related Work}
\label{related}
%!TEX root = ms.tex
% \vspace{-0.05 in}
\paragraph{Single-Agent Exploration} In order to solve sparse reward problems, researchers have long worked on improving exploration in reinforcement learning.
Prior works commonly propose reward bonuses that encourage agents to reach novel states.
In tabular domains, reward bonuses based on the inverse state-action count have been shown to be effective in accelerating learning~\citep{STREHL20081309}.
In order to scale count-based approaches to large state spaces, many recent works have focused on devising pseudo state counts to use as reward bonuses~\citep{bellemare2016unifying,ostrovski2017count,tang2017exploration}.
Alternatively, some work has focused on defining intrinsic rewards for exploration based on inspiration from psychology~\citep{oudeyer2009intrinsic,schmidhuber2010formal}.
These works use various measures of state novelty as intrinsic rewards motivating exploration~\citep{pmlr-v70-pathak17a,houthooft2016vime,burda2018exploration}

% \vspace{-0.04 in}
% \paragraph{Multi-Agent Reinforcement Learning (MARL)} Multi-agent reinforcement learning introduces several unique challenges that recent work attempts to address.
% These challenges include: multi-agent credit assignment in cooperative tasks with shared rewards~\citep{sunehag2017value,rashid2018qmix,Foerster2017-do}, non-stationarity of the environment in the presence of other learning agents~\citep{lowe2017multi,Foerster2017-do,pmlr-v97-iqbal19a}, and learning of communication protocols between cooperative agents~\citep{foerster2016learning,sukhbaatar2016learning,jiang2018learning}.

% \vspace{-0.04 in}
\paragraph{Exploration in MARL}
% ~\citet{carmel1997exploration} consider exploration with respect to opponent strategies in competitive games, and~\citet{verbeeck2005coordinated} consider exploration of a large joint action space in a load balancing problem.
~\citet{Jaques2018-gu} define an intrinsic reward function for multi-agent reinforcement learning that encourages agents to take actions which have the biggest effect on other agents' behavior, otherwise referred to as ``social influence''.
~\citet{agogino2008analyzing} defines metrics for evaluating the efficacy of reward functions in multi-agent domains.
These works, while important, do not address the problem of coordinating exploration in a large state space among multiple agents.
A recent approach to collaborative evolutionary reinforcement learning~\citep{khadka2019collaborative} shares some similarities with our approach.
As in our work, the authors devise a method for learning a population of diverse policies with a shared replay buffer and dynamically selecting the best learner; however, their work is focused on single-agent tasks and does not incorporate any notion of intrinsic rewards for exploration.
\citet{wang2020influencebased} define influence-based rewards which encourage agents to visit regions where their actions \emph{influence} other agents' transitions and rewards (e.g. one agent unlocks a door for another).
In practice, they combine their approach with state-based exploration, and thus, their work is complementary and orthogonal to our approach where agents do not simply explore regions that are novel to them but take into account how novel all other agents consider these regions.
This work is most applicable to settings where agents influencing each others' dynamics forms a positive inductive bias towards solving a task.
Most recently, \citet{mahajan2019maven} introduce a mechanism for achieving ``committed'' exploration, allowing agents to explore temporally extended coordinated strategies.
While this approach enables coordinated exploration, it does not encourage exploration of novel states, and as such, may not learn effectively in sparse reward tasks.

%!TEX root = ms.tex
\section{Background}
\label{background}

\paragraph{Dec-POMDPs} In this work, we consider the setting of decentralized POMDPs~\citep{oliehoek2016concise}, which are used to describe cooperative multi-agent tasks.
A decentralized POMDP (Dec-POMDP) is defined by a tuple: $(\mathbf{S}, \mathbf{A}, \mathit{T}, \mathbf{O}, \mathit{O}, \mathit{R}, n, \gamma)$.
In this setting we have $n$ total agents.
$\mathbf{S}$ is the set of global states in the environment, while $\mathbf{O} = \otimes_{i \in \{ 1 \dots n \}} \mathbf{O}_i$ is the set of joint observations for each agent and $\mathbf{A} = \otimes_{i \in \{ 1 \dots n \}} \mathbf{A}_i$ is the set of possible joint actions for each agent.
A specific joint action at one time step is denoted as $\mathbf{a} = \{a_1, \dots, a_n \} \in \mathbf{A}$ and a joint observation is $\mathbf{o} = \{o_1, \dots, o_n \} \in \mathbf{O}$.
$\mathit{T}$ is the state transition function which defines the probability $P(s' | s, \mathbf{a})$, and $\mathit{O}$ is the observation function which defines the probability $P(\mathbf{o} | \mathbf{a}, s')$.
$\mathit{R}$ is the reward function which maps the combination of state and joint actions to a single scalar reward.
Importantly, this reward is shared between all agents, so Dec-POMDPs always describe cooperative problems.
Finally, $\gamma$ is the discount factor which determines how much the agents should favor immediate reward over long-term gain.

\paragraph{Soft Actor-Critic}
Our approach uses Soft Actor-Critic (SAC)~\citep{haarnoja2018soft} as its underlying algorithm.
SAC incorporates an entropy term in the loss functions for both the actor and critic, in order to encourage exploration and prevent premature convergence to a sub-optimal deterministic policy.
The policy gradient with an entropy term is computed as follows:

\vspace{-0.25 in}
\begin{multline}
    \nabla_\theta J(\pi_\theta) = \mathds{E}_{s \sim D, a \sim \pi} \biggl[ \nabla_\theta \log \pi_\theta(a | s) \biggl(-\frac{\log \pi_\theta(a | s)}{\alpha} +  \\
    Q_\psi(s, a) - b(s) \biggr) \biggr]
\end{multline}
where $D$ is a replay buffer that stores past environment transitions, $\psi$ are the parameters of the learned critic, $b(s)$ is a state dependent baseline (e.g. the state value function $V(s)$), and $\alpha$ is a reward scale parameter determining the amount of entropy in an optimal policy.
The critic is learned with the following loss function:

\vspace{-0.15 in}
\begin{equation}
\label{eq:sac_qloss}
    \mathcal{L}_Q(\psi) = \mathds{E}_{(s, a, r, s') \sim D} \left[ (Q_\psi(s,a) - y)^2 \right]
\end{equation}
\vspace{-0.15 in}
\begin{equation}
    y = r(s,a) + \gamma \mathds{E}_{a' \sim \pi} \left[Q_{\bar{\psi}}(s', a') - \frac{\log(\pi_{\bar{\theta}}(a' | s'))}{\alpha} \right]
\end{equation}
where $\bar{\psi}$ are the parameters of the target critic which is an exponential moving average of the past critics, updated as: $\bar{\psi} \leftarrow (1 - \tau) \bar{\psi} + \tau \psi$, and $\tau$ is a hyperparameter that controls the update rate.

\paragraph{Centralized Training with Decentralized Execution}
A number of works in deep multi-agent reinforcement learning have followed the paradigm of centralized training with decentralized execution~\citep{lowe2017multi,Foerster2017-do,sunehag2017value,rashid2018qmix,pmlr-v97-iqbal19a}.
This paradigm allows for agents to train while sharing information (or incorporating information that is unavailable at test time) but act using only local information, without requiring communication which may be costly upon execution.
Since most reinforcement learning applications use simulation for training, communication between agents during the training phase has a relatively low cost.

\section{Intrinsic Rewards for MARL Exploration}
\label{sec:intr_rews}
In this section we describe intrinsic reward functions for exploration that are specifically tailored to multi-agent learning.
The main idea is to share whether other agents have explored a region and consider it as novel.

We assume that each agent (indexed by $i$) has a novelty function $f_i: \mathbf{O}_i \rightarrow \mathbb{R}^+$ that determines how novel an observation is to it, based on its past experience.
This function can be an inverse state visit count in discrete domains, or, in large/continuous domains, it can be represented by recent approaches for developing novelty-based intrinsic rewards in complex domains, such as random network distillation~\citep{burda2018exploration}.
We assume that all agents share the same observation space so that each agent's novelty function can operate on all other agents' observations.

We define the multi-agent intrinsic reward function  $g_i(\cdot)$ for agent $i$, which considers how novel \emph{all} agents consider $i$'s observation. Concretely, the function maps the vector $[f_1(o_i),~\cdots,~f_n(o_i)]$ to a scalar reward.

\paragraph{Desiderata of $g_i(\cdot)$} While in theory $g_i(\cdot)$ can be in any form, we believe the following two properties are intuitive and naturally applicable to cooperative MARL:

\begin{itemize}
     \item \textbf{Coordinate-wise  Monotonicity}\ An observation becoming less novel to any individual agent should \emph{not increase} the intrinsic reward, preventing the agent from exploring a region more as it becomes more known to another agent. Formally,  $\partial g_i/\partial f_j \geq 0, \forall~i,j$. 
     \item \textbf{Inner-directedness}\ If an observation approaches having zero novelty to an agent, then the intrinsic reward should also approach zero, irrespective of other agents' novelty.
     % In other words, the intrinsic reward function can always be written as $g_i(\cdot) = f_i(o_i)g_i'(\cdot)$.
     This prevents the agent from repetitively exploring the same regions, at the ``persuasion'' of other agents.
\end{itemize}

\begin{figure}[t]
    \centering
    \includegraphics[width=0.8\linewidth]{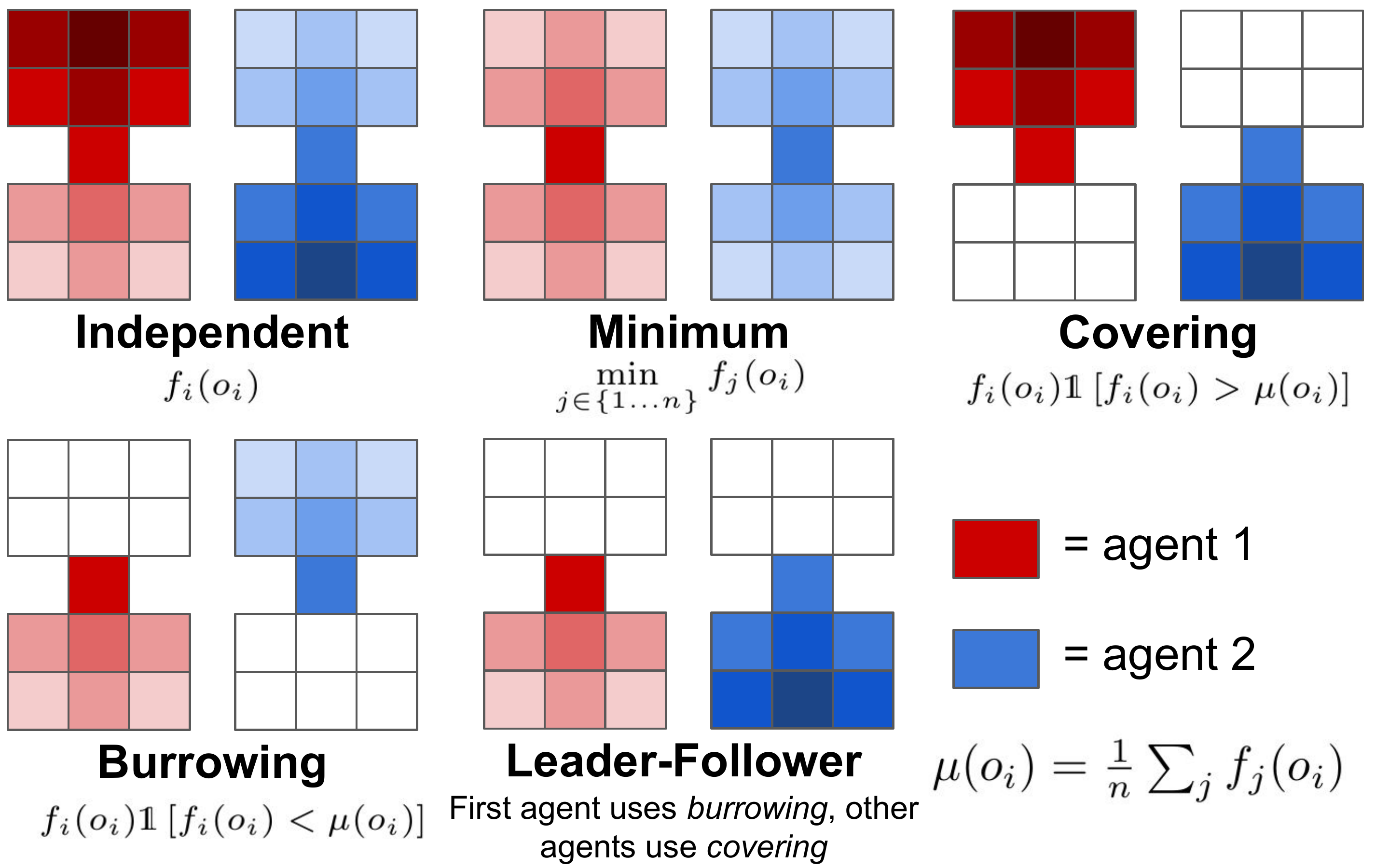}
    \caption{\small Multi-agent intrinsic rewards. Visualized for the 2 agent case. \textsc{Independent} shows the regions that have been explored by each agent. Darker shades mean higher reward values.}
    \label{fig:intr_rews}
\end{figure}

\paragraph{Examples} Fig.~\ref{fig:intr_rews} visualizes examples of intrinsic rewards that observe the aforementioned desirable properties. \textsc{Independent} rewards are analagous to single-agent approaches to exploration which define the intrinsic reward for an agent as the novelty of their own observation that occurs as a result of an action.
The remainder of intrinsic reward functions that we consider use the novelty functions of other agents, in addition to their own, to further inform their exploration.

\textsc{Minimum} rewards consider how novel all agents find a specific agent's observation and rewards that agent based on the minimum.
This method leads to agents only being rewarded for exploring areas that no other agent has explored, which could be advantageous in scenarios where redundancy in exploration is not useful or even harmful.
\textsc{Covering} rewards agents for exploring areas that it considers more novel than the average agent.
This reward results in agents shifting around the state space, only exploring regions as long as they are more novel to them than their average teammate.
\textsc{Burrowing} rewards do the opposite, only rewarding agents for exploring areas that it considers less novel than the average agent.
While seemingly counterintuitive, these rewards encourage agents to further explore areas they have already explored with the hope that they will discover new regions that few or no other agents have seen, which they will then consider less novel than average and continue to explore.
As such, these rewards result in agents continuing to explore until they exhaust all possible intrinsic rewards from a given region (i.e. hit a dead end), somewhat akin to a depth-first search.
\textsc{leader-follower} uses burrowing rewards for the first agent, and covering rewards for the rest of the agents. This leads to an agent exploring a space thoroughly, and the rest of the agents following along and trying to cover that space.

Note that these are not meant to be a comprehensive set of intrinsic reward functions applicable to all cooperative multi-agent tasks.
In fact, any convex combination of them is consistent with the desiderata.
We use the set of rewards introduced in Fig.~\ref{fig:intr_rews} to concurrently learn a set of diverse exploration policies and introduce in the next section a method for dynamically switching between them.
This method matches or exceeds the performance of the best individual exploration method in several complex tasks.

\section{Learning for Multi-Agent Exploration}

For many tasks, it is impossible to know a priori what type of exploration may provide the right inductive bias.
Furthermore, the type that is most helpful could change over the course of training if the task is sufficiently complex.
In this section we present our approach for simultaneously learning policies trained with different types of intrinsic rewards and dynamically selecting the best one.

\begin{figure}[t]
    \centering
    \includegraphics[width=0.7\linewidth]{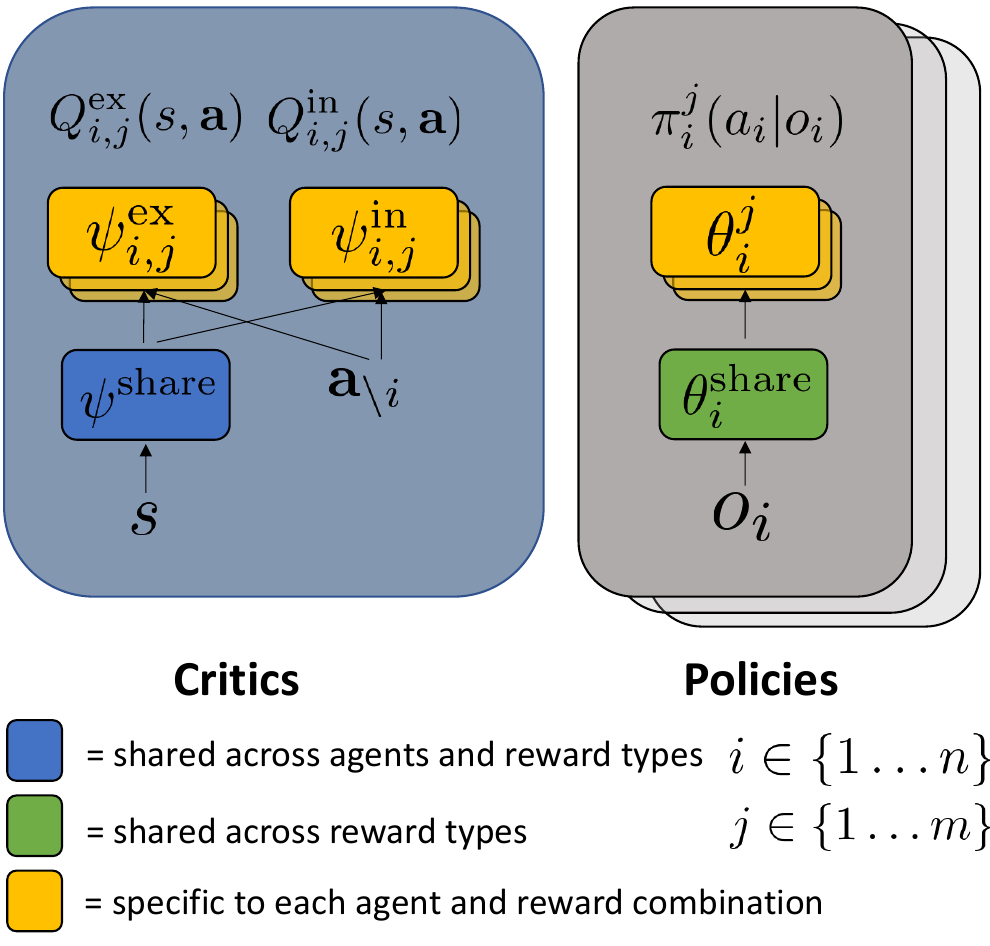}
    \caption{\small Diagram of our model architecture. Colors indicate how parameters are shared. $i$ indexes agents, while $j$ indexes reward types. Full architectures in pseudo-code are provided in the Supp.}
    \label{fig:model_diagram}
    \vspace{-0.15 in}
\end{figure}

\paragraph{Simultaneous Policy Learning}
In order to learn policies for various types of intrinsic rewards in parallel, we utilize a shared replay buffer and off-policy learning to maximize sample efficiency.
In other words, we learn policies and value functions for \emph{all} intrinsic reward types from \emph{all} collected data, regardless of which policies it was collected by.
This parallel learning is made possible by the fact that we can compute our novelty functions off-policy, given the observations for each agent after each environment transition, which are saved in a replay buffer.
In Figure~\ref{fig:model_diagram} we visualize our model architecture.
We share a critic base and split extrinsic and intrinsic return heads as in~\citet{burda2018exploration}.
We learn separate heads for each agent $i \in \{ 1 \dots n \}$ and reward $j \in \{ 1 \dots m \}$ where $m$ is the total number of intrinsic reward types that we are considering.
For policies, each agent learns its own base that is shared across separate heads for all intrinsic reward types.
Our specific learning algorithm is adapted from the multi-agent Soft-Actor-Critic method presented in~\citet{pmlr-v97-iqbal19a}.

The policy for agent $i$, trained using reward $j$ (in addition to extrinsic rewards), is represented by $\pi_i^j$.
% It takes as input agent $i$'s observation, $o_i$, and outputs a distribution from which we can sample the action $a_i$.
The parameters of this policy are $\Theta_i^j = \{\theta^\text{share}_i, \theta_i^j \}$, where $\theta^\text{share}_i$ is the shared base/input (for agent $i$) and $\theta_i^j$ is a head/output specific to this reward type. The extrinsic critic for policy head $\pi_i^j$ is represented by $Q^\text{ex}_{i,j}$.
It takes as input the global state $s$ and the actions of all other agents $\mathbf{a}_{\setminus i}$, and it outputs the expected returns under policy $\pi_i^j$ for each possible action that agent $i$ can take, given all other agents' actions.
The parameters of this critic are $\Psi^\text{ex}_{i,j} = \{ \psi^\text{share}, \psi^\text{ex}_{i,j} \}$ where $\psi^\text{share}$ is a shared base across all agents and reward types.
A critic with similar structure exists for predicting the intrinsic returns of actions taken by $\pi^j_i$, represented by $Q^\text{in}_{i,j}$, which uses the parameters: $\Psi^\text{in}_{i,j} = \{ \psi^\text{share}, \psi^\text{in}_{i,j} \}$.
Note that the intrinsic critics share the same base parameters $\psi^\text{share}$.

% We remove the symbols representing the parameters of the policies ($\Theta$) and the critics ($\Psi$) for readability.
In our notation, we use the absence of a subscript or superscript to refer to a group.
For example $\pi^j$, refers to \emph{all} agents' policies trained on intrinsic reward $j$.
We train our critics with the soft actor-critic Q-function loss (Equation~\ref{eq:sac_qloss}), using the following target, $y^\text{ex/in}_{i,j}$, for the extrinsic and intrinsic critics:

\vspace{-0.2 in}
\begin{equation}
    r^\text{ex/in} +
    \gamma \mathds{E}_{\mathbf{a}' \sim \bar{\pi}^j} \left[ \bar{Q}^\text{ex/in}_{i,j}(s', \mathbf{a}') - \frac{\log(\bar{\pi}^j_i(a_i' | o_i'))}{\alpha} \right]
\end{equation}
The extrinsic target, $y^\text{ex}_{i,j}$, uses the Dec-POMDP's reward function $r^\text{ex}(s,\mathbf{a})$, while the intrinsic target, $y^\text{in}_{i,j}$ uses the intrinsic reward $r^\text{in}_{i,j}(o_i')$.
Note that the intrinsic rewards depend on the observation resulting from the actions taken, $o_i'$.
$\bar{Q}^\text{ex/in}_{i,j}$ refers to the target Q-function, an exponential weighted average of the past Q-functions, used for stability.
$\bar{\pi}^j_i$ are similarly updated target policies.
We train each policy head with the following gradient:

\vspace{-0.15 in}
\begin{multline}
    \nabla_{\Theta^j_i} J(\pi^j_i) = \mathds{E}_{(s, \mathbf{o}) \sim D, \mathbf{a} \sim \pi^j} \Bigl[ \nabla_{\Theta^j_i} \log \pi^j_i(a_i | o_i) \\
    \left(-\frac{\log \pi_i^j(a_i | o_i)}{\alpha} + A^j_i(s, \mathbf{a})\right) \Biggr]
\end{multline}
\vspace{-0.1 in}
\begin{equation}
    A^j_i(s, \mathbf{a}) = Q^\text{ex}_{i,j}(s, \mathbf{a}) + \beta Q^\text{in}_{i,j}(s, \mathbf{a}) - V^j_i(s, \mathbf{a}_{\setminus i})
\end{equation}
\vspace{-0.15 in}
\begin{multline}
    V^j_i(s, \mathbf{a}_{\setminus i}) = \sum_{a_i' \in \mathbf{A}_i} \pi_i^j(a_i'|o_i) (Q^\text{ex}_{i,j}(s, \{a_i', \mathbf{a}_{\setminus i} \}) + \\
    \beta Q^\text{in}_{i,j}(s, \{a_i', \mathbf{a}_{\setminus i} \}))
\end{multline}
where $\beta$ is a scalar that determines the weight of the intrinsic rewards, relative to extrinsic rewards, and $A^j_i$ is a multi-agent advantage function~\citep{Foerster2017-do,pmlr-v97-iqbal19a}, used for helping with multi-agent credit assignment.
We update \emph{all} policy and critic heads with each environment transition sample.

\paragraph{Dynamic Policy Selection}
\label{pol-sel}

Now that we have established a method for simultaneously learning policies using different intrinsic reward types, we must devise a means of selecting between these policies when collecting data in the environment (i.e., rollouts).
In order to select policies to use for rollouts, we must consider which policies maximize extrinsic returns, while taking into account the fact that there may still be ``unknown unknowns,'' or regions that the agents have not seen yet where they may be able to further increase their extrinsic returns.
As such, we must learn a meta-policy that selects from the set of policies trained on different intrinsic rewards to maximize extrinsic returns while maintaining some degree of stochasticity.
We parameterize the selector policy $\Pi$ with a vector, $\phi$, that contains an entry for every reward type.
The probability of sampling head $j$ is: $\Pi(j) \propto \exp(\phi[j])$.
We sample from the meta-policy at the start of each rollout.
Using policy gradients, we train the policy selector, $\Pi$, to maximize extrinsic returns:
\vspace{-0.05 in}
\begin{multline}
    \nabla_\phi J(\Pi) = \mathds{E}_{h \sim \Pi} \Bigl[ \nabla_\phi \log \Pi(h) \\
    \left( -\frac{\log \Pi(h)}{\eta} + R^\text{ex}_h - b_\Pi \right) \Bigr]
\end{multline}
\vspace{-0.15 in}
\begin{equation}
    R^\text{ex}_h = \sum_{t=0}^T \gamma^t r^\text{ex}(s_t, \mathbf{a}_t) | \mathbf{a} \sim \pi^h,~b_\Pi = \sum_{h'}^m \Pi(h') \mu_{h'}
\end{equation}
$\mu_h$ is a running mean of the returns received by head $h$ in the past, and $\eta$ is a parameter similar to $\alpha$ for the low-level policies, which promotes entropy in the selector policy.
Entropy in the policy selector prevents it from collapsing onto a single exploration type that exploits a local optimum.
% This training procedure encourages the policy selector to select policies that have the highest expected returns from the agents' initial positions.
% As such, we can learn a diverse set of behaviors based on various multi-agent intrinsic reward functions and select the one that maximizes performance on the task at hand at any point during training, while continuing to consider other policies that may lead to greater rewards.
% The hierarchical approach we describe allows for diverse coordinated exploration behaviors that succeed in a variety of settings, as we show in Section~\ref{experiments}.

\section{Experiments}
\label{experiments}
%!TEX root = ms.tex

\begin{figure*}[t]
    \centering
    \begin{subfigure}[t]{0.2\linewidth}
        \centerline{\includegraphics[width=\linewidth]{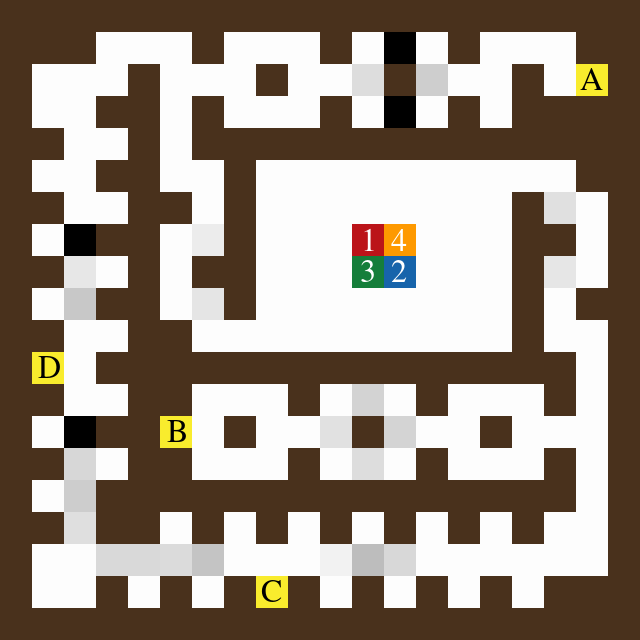}}
        \caption{}
        \label{fig:env_render}
    \end{subfigure}
    \hfill
    \begin{subfigure}[t]{0.2\linewidth}
        \centerline{\includegraphics[width=\linewidth]{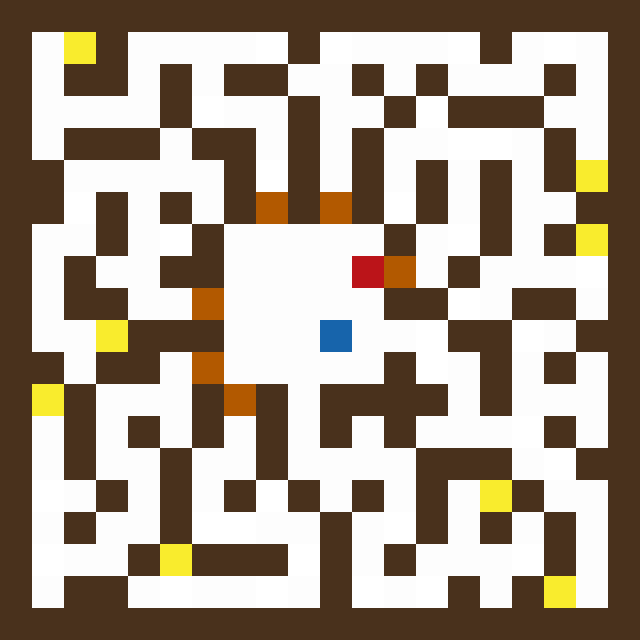}}
        \caption{}
        \label{fig:gridworld_tiered}
    \end{subfigure}
    \hfill
    \begin{subfigure}[t]{0.223\linewidth}
        \centerline{\includegraphics[width=\linewidth]{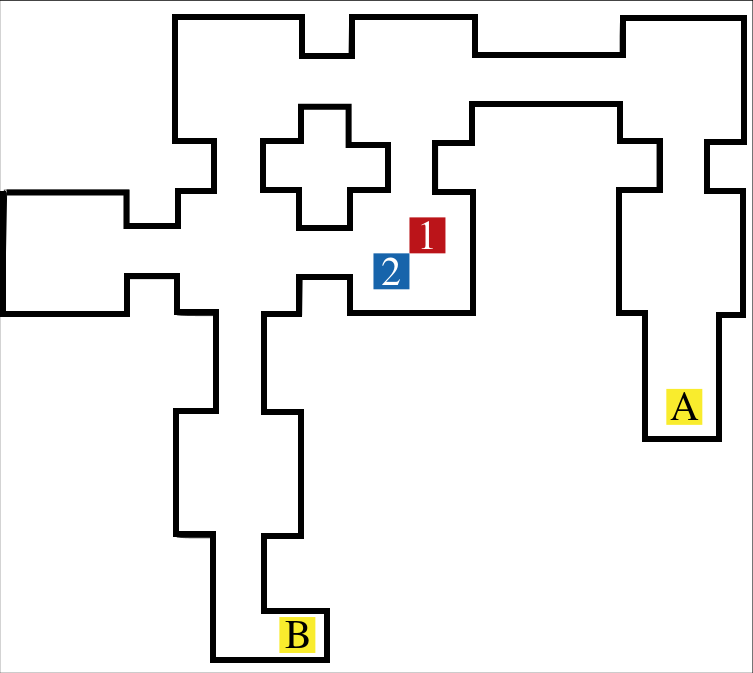}}
        \caption{}
        \label{fig:vzd_topdown}
    \end{subfigure}
    \hfill
    \begin{subfigure}[t]{0.27\linewidth}
        \centerline{\includegraphics[width=\linewidth]{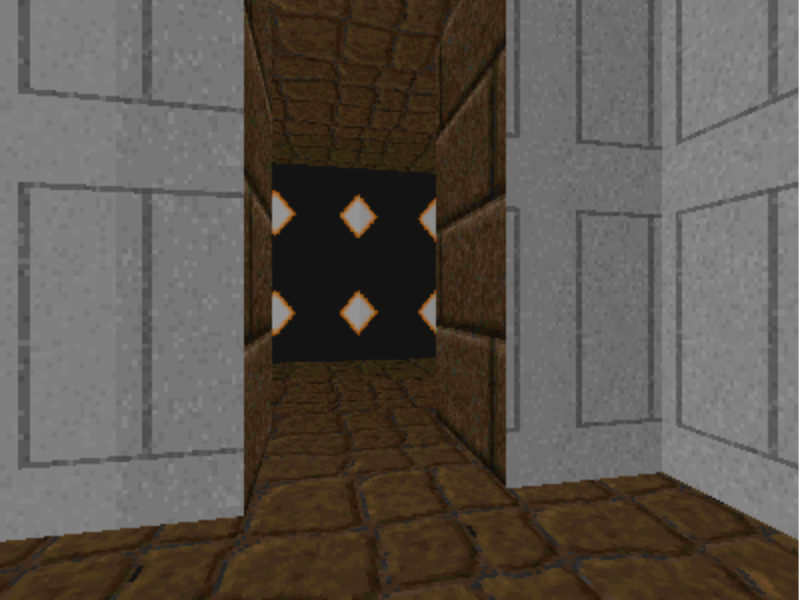}}
        \caption{}
        \label{fig:vzd_egocentric}
    \end{subfigure}
    \caption{\small (a) Rendering of the fixed map in our gridworld domain.
                    % Agents start each episode in the central room and must complete various tasks related to collecting the yellow treasures placed around the map.
                    (b) Randomly generated map used for \textbf{\textsc{Flip-Task}} task.
                    $n$ doors to new paths open every time either \textbf{\textsc{task 1}} or \textbf{\textsc{task 2}} is completed.
                    (c) Top-Down view of VizDoom ``My Way Home'' map, modified for multi-agent experiments.
                    (d) Egocentric view in VizDoom used for agents' observations.
        }
    \label{fig:all_envs}
    \vspace{-0.15 in}
\end{figure*}
We first train policies with each intrinsic reward function defined in Fig.~\ref{fig:intr_rews}, then compare our approach as well as several baselines and ablations.
We refer to the best performing reward type for each setting as the ``non-adaptive oracle''.
``Non-adaptive'' is used to contrast with our approach which can adapt different exploration strategies during training, while ``oracle'' is used since we do not know a priori which type will perform best.
In our experiments we validate the following hypotheses: 1) Multi-agent intrinsic reward functions improve performance on tasks requiring coordination, 2) Our approach matches the performance of the non-adaptive oracle without training separate policies, and 3) In tasks requiring changing coordination strategies, our method outperforms the non-adaptive oracle.
% We begin by describing our evaluation tasks and domains, followed by experimental results.
% We provide additional details in the supplementary material as well as code for both the algorithm and environments.

\subsection{Tasks}
Tasks for testing single-agent exploration typically revolve around navigation of an environment with sparsely distributed rewards (e.g. Montezuma's Revenge~\citep{ostrovski2017count,tang2017exploration,burda2018exploration}, VizDoom~\citep{pmlr-v70-pathak17a}, etc).
In the multi-agent setting, we define tasks which similarly consider navigation with very sparse rewards, while requiring varying modalities of coordination across agents.
These tasks involve collecting the items spread around a map (displayed in yellow in Figure~\ref{fig:all_envs}):

\textbf{\textsc{task 1}:}  Agents must cooperatively collect \emph{all} treasure on the map in order to complete the task.
Ideally, agents should spread out in order to solve the task effectively.
\textbf{\textsc{task 2}:} Agents must all collect the \emph{same} treasure. Thus, agents would ideally explore similar regions concurrently.
\textbf{\textsc{task 3}:} Agents must all collect the specific treasure assigned to them, requiring no coordination across agents.

\textbf{\textsc{task 1}} and \textbf{\textsc{task 2}} need to solve coordination problems, as an individual agent can repeat the same behavior and receive drastically different returns depending on the behavior of other agents.
\textbf{\textsc{task 3}} is intended as a sanity check where independent exploration should perform best.
All 3 tasks are tested on the maps pictured in Figures~\ref{fig:env_render} and~\ref{fig:vzd_topdown}.

\textbf{\textsc{Flip-Task}} is a task where the modality of required coordination changes as agents progress, akin to the search and rescue task mentioned in the introduction.
This task is tested on randomly generated maps, an example of which is pictured in Figure~\ref{fig:gridworld_tiered}.
In \textbf{\textsc{Flip-Task}} agents begin in a central room with $n$ branching paths available (where treasures are placed at the furthest available point) and must solve either \textbf{\textsc{task 1}} or \textbf{\textsc{task 2}} with respect to the available treasures.
Once this task is complete, the next set of $n$ paths (blocked by the light brown doors) opens up for which the task will be the opposite of the previous task (1 $\rightarrow$ 2, 2 $\rightarrow$ 1), requiring agents to adapt their exploration strategy after they learn to solve the first task.

% The two agent version of each task uses agents 1-2 and treasure A-B, while the three agent versions use 1-3, A-C, and the four agent versions use 1-4, A-D.
Agents receive a negative time penalty in their extrinsic rewards at each step, so they are motivated to complete the task as quickly as possible.
The only positive extrinsic reward comes from any agent collecting a treasure allowed by the specific task, and rewards are shared between all agents.

\vspace{-0.07 in}
\subsection{Domains}
We first test our approach using a multi-agent gridworld domain (pictured in Fig.~\ref{fig:env_render} and~\ref{fig:gridworld_tiered}).
% At each step there is a 10\% chance of an agent's action being replaced by a random one.
% Inspired by the enemies in Montezuma's Revenge which increase the challenge of exploration, we introduce ``wormholes'' which have a probability of opening at each time step.
% This probability, which agents can observe, changes at each step using a biased random walk such that it moves toward one, until the hole opens and it resets to zero.
% If an agent steps into an open wormhole, they will be sent back to their starting position.
Then, in order to test our method's ability to scale to more complex 3D environments with visual observations, we test on the VizDoom framework~\citep{Kempka2016ViZDoom}.

The novelty function for each agent $f_i$, which is used for calculating the intrinsic rewards in Figure~\ref{fig:intr_rews}, is defined as $\frac{1}{N^{\zeta}}$, where $N$ is the number of times the agent has visited its current cell and $\zeta$ is a decay rate selected as a hyperparameter (we find $\zeta = 0.7$ works well for our purposes).
Since VizDoom is not a discrete domain, we discretize agents' $(x,y)$ positions into bins and use the counts for these bins.

\vspace{-0.07 in}
\subsection{Baselines and Ablations}
We consider another approach to adapting single-agent intrinsic rewards to MARL in \textit{Centralized}, where we provide intrinsic rewards to all agents as if they were a single agent.
In other words, we use the inverse count of the number of times \textbf{all} agents have jointly taken up their combined positions.
% Our other baselines consist of state-of-the-art cooperative multi-agent RL algorithms.
We also evaluate QMIX~\citep{rashid2018qmix}, a state-of-the-art method in cooperative MARL, and MAVEN~\citep{mahajan2019maven}, which builds on QMIX by incorporating committed temporally extended exploration.
We use the authors' open-sourced code for these comparisons.

Finally, we conduct ablation studies to determine the effectiveness of our meta-policy in balancing exploration and exploitation.
\textit{Multi (Uniform Meta-Policy)} samples action policies at random and \textit{Multi (No Entropy)} does not incorporate entropy to encourage exploration.
We also tested the base multi-agent SAC algorithm without intrinsic rewards (MA-SAC).

\vspace{-0.07 in}
\subsection{Results}
\begin{figure}[ht]
    \centering
    \centerline{\includegraphics[width=\linewidth]{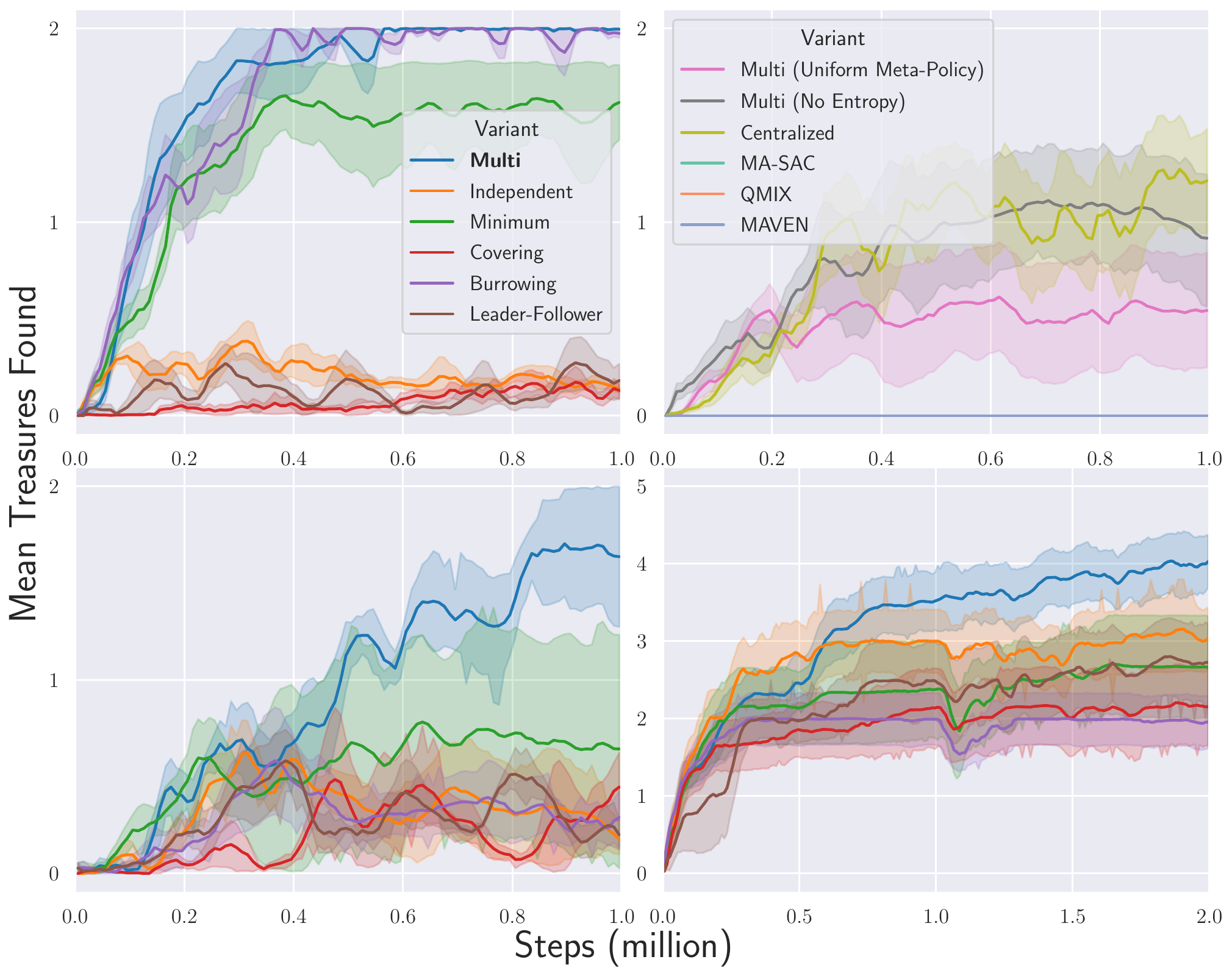}}
    \caption{
        \small
        (Top Left) Mean number of treasures found per episode on \textbf{\textsc{task 1}} with 2 agents in the gridworld domain.
         % Our approach (\textsc{Multi}) is competitive with the non-adaptive oracle (\textsc{Burrowing}), using the same number of environment samples without any prior knowledge provided.
        (Top Right) Ablations and baselines in the same setting.
        (Bottom Left) \textbf{\textsc{task 3}} with 2 agents on VizDoom.
        (Bottom Right) \textbf{\textsc{Flip-Task}} with 2 agents on gridworld.
        % We show both aspects of our approach (the meta-policy selector and the diverse intrinsic reward functions) are crucial for successful completion of exploration tasks requiring coordination.
        % Furthermore, we show that the intrinsic rewards developed within our framework are advantageous when compared to a naive centralized approach to exploration.
        Shaded regions are a 68\% confidence interval across 6 runs of the running mean over the past 100 episodes.
        }
    \label{fig:comp_results}
    \vspace{-0.15 in}
\end{figure}

Results (non-adaptive oracle and our approach) in all settings are summarized in Table~\ref{tab:all_task_res}, and several training curves are found in Figure~\ref{fig:comp_results}.
The detailed training curves and a table of final performance for all intrinsic reward types in all settings can be found in the supplementary material.
We note the type of the non-adaptive oracle is frequently not \textsc{independent}, indicating that the multi-agent intrinsic rewards introduced in section~\ref{sec:intr_rews} are effective in comparison to a naive application of single agent intrisic reward methods.
We also include in the supplement a discussion of the behavior of each individual reward type along with videos demonstrating this behavior.

\begin{table}[t]
\captionsetup{width=1.0\linewidth,justification=centering}
\begin{center}
    \scriptsize
    \begin{tabular}{@{}lllll@{}}
        \toprule
        \multicolumn{5}{c}{\textbf{\textsc{Gridworld}}} \\
        \midrule
        Task & $n$ & \textsc{NAO} Type & \textsc{NAO} & \textbf{\textsc{Multi}} \\
        \midrule
        1 & 2 & \textsc{Burrowing} & {1.98 $\pm$ 0.06} & {2.00 $\pm$ 0.00} \\
          & 3 & \textsc{Burrowing} & {2.06 $\pm$ 1.05} & {2.23 $\pm$ 0.73} \\
          & 4 & \textsc{Burrowing} & {1.90 $\pm$ 0.49} & {2.04 $\pm$ 0.61 }\\
        \midrule
        2 & 2 & \textsc{Independent} & {2.00 $\pm$ 0.00} & {1.83 $\pm$ 0.41} \\
          & 3 & \textsc{Lead-Follow} & {3.00 $\pm$ 0.00} & 1.80 $\pm$ 0.71 \\
          & 4 & \textsc{Lead-Follow} & {2.66 $\pm$ 2.06} & {2.54 $\pm$ 1.21} \\
        \midrule
        3 & 2 & \textsc{Independent} & {1.39 $\pm$ 0.94} & {2.00 $\pm$ 0.00} \\
          & 3 & \textsc{Independent} & {1.68 $\pm$ 0.70} & {2.21 $\pm$ 0.91} \\
          & 4 & \textsc{Burrowing} & {2.14 $\pm$ 1.49} & {1.73 $\pm$ 0.47} \\
        \midrule
        Flip-Task & 2 & \textsc{Independent} & {3.04 $\pm$ 1.33} & {4.03 $\pm$ 0.97} \\
        \midrule
        \multicolumn{5}{c}{\textbf{\textsc{VizDoom}}} \\
        \midrule
        1 & 2 & \textsc{Burrowing} & {1.94 $\pm$ 0.10} & {1.98 $\pm$ 0.03} \\
        2 & 2  & \textsc{Lead-Follow} & {1.93 $\pm$ 0.10} & {1.23 $\pm$ 0.65} \\
        3 & 2  & \textsc{Minimum} & {0.64 $\pm$ 1.05} & {1.64 $\pm$ 0.63 } \\
        \bottomrule
    \end{tabular}
    \caption{\small \# of treasures found  with standard deviation across 6 runs. Our method (\textbf{\textsc{multi}}) matches or outperforms the non-adaptive oracle (\textsc{NAO}) in nearly all settings}
    \label{tab:all_task_res}
    \vspace{-0.2 in}
\end{center}
\end{table}

% \paragraph{Do the Oracle Types Align with the Tasks?}
% We found the reward types of the non-adaptive oracles to largely align with the intuitions behind the design of the tasks.
% \textsc{independent} rewards, as expected, result in agents exploring the whole state space without taking other agents into consideration.
% Interestingly, policies trained with these rewards can fail on~\textbf{\textsc{Task 3}}, which assigns independent goals to each agent and does not require coordination.
% Since our maps contain branching non-linear paths, we find that these agents occasionally suffer from detachment~\citep{ecoffet2019go} (ie. forgetting how to return to an previously explored area).
% Both \textsc{minimum} and \textsc{burrowing} encourage agents to not explore the same regions redundantly and are thus successful on \textbf{\textsc{Task 1}} which requires divide-and-conquer type behavior.
% \textsc{Leader-Follower} encourages the opposite behavior and is thus most successful on \textbf{\textsc{Task 2}}.
% \textsc{covering} rewards, as expected, lead to behavior where agents are constantly switching the regions they explore.
% While this behavior alone does not prove to be useful in the tasks we test since the switching slows down overall exploration progress, it serves as a useful mechanism to break out of bad local optima for our meta-policy approach.
% The unique behavior of each reward type is visualized in videos included in the supplementary material.

\paragraph{Matching the Oracle Performance with a Meta-Policy}
We find our approach is competitive with the non-adaptive oracle in nearly all tasks, while only needing the same number of samples as a \emph{single} run (a more fair comparison would allow our method to train for the sum of samples used to identify the oracle).
This performance is exciting, as our method receives no prior information about the optimal type of exploration, while each type carries its own bias.
Furthermore, we find our results on the more complex VizDoom domain mirror those in the gridworld, indicating our methods are not limited to discrete domains, assuming a reliable way for measuring the novelty of observations exists.

\paragraph{Advantages of Adaptive Strategies}
In two cases we find our approach is able to surpass the performance of the non-adaptive oracle: ~\textbf{\textsc{Task 3}} and \textbf{\textsc{Flip-Task}}.
In the case of \textbf{\textsc{Task 3}}, this is interesting as rewards for each agent are assigned independently, so coordination is not strictly necessary.
In this case, multi-agent intrinsic rewards introduce useful biases that naturally divide the space into explorable regions for each agent, reducing the chances of "detachment" ~\citep{ecoffet2019go} (which we find independent exploration to suffer from); however, in this task they may divide the space into the \emph{wrong} regions for each agent to get their rewards.
Our approach succeeds due to its ability to reap the benefits of multi-agent intrinsic rewards while being able to switch strategies if one is not working.
We analyze this meta-policy behavior in detail in the supplementary material.
The success on \textbf{\textsc{Task 3}} suggests that our approach may be beneficial in the single-agent setting when potential paths of exloration are numerous (e.g. by training several policies for one task and treating them as separate agents in our reward functions).

Our approach is similarly successful on \textbf{\textsc{Flip-Task}}.
Due to the adaptability of our meta-policy, the agents are able to switch their exploration type to suit the next task after they learn to solve the first one.
Note that all sets of exploration policies will learn to solve the first task once any single one does since they share experience and are trained on a combination of extrinsic and intrinsic rewards.
As such, the meta-policy can switch to an exploration strategy best suited for the next task, knowing that it will reliably solve the previous task.

\paragraph{The Unexpected Challenge of \textsc{task 2}}
We find our approach is unable to match the performance of the non-adaptive oracle on \textbf{\textsc{Task 2}} in certain cases (gridworld with 3 agents and VizDoom).
This lack of success may be an indication that the exploration strategies which perform well in these settings require commitment to a single strategy early on in training, highlighting a limitation of our approach.
Our method requires testing out all policies until we find one that reaches high extrinsic rewards, which can dilute the effectiveness of exploration early on.

\paragraph{Ablations and Baselines} 
We test against several baselines in the gridworld setting on \textbf{\textsc{Task 1}} with 2 agents (top right of Fig.~\ref{fig:comp_results}).
Training curves for other tasks can be found in the Supplement.
We find that those results show the same patterns as \textbf{\textsc{Task 1}}.
We find that our approach balances meta-exploration and exploitation by outperforming both \textit{Multi (Uniform Meta-Policy)} (pure explore) and \textit{Multi (No Entropy)} (pure exploit).
While \textit{Centralized} will ensure the global state space is thoroughly searched, it lacks the inductive biases toward spatial coordination that our reward functions incorporate.
As such, it does not learn as efficiently as our method.
Finally, all three of MA-SAC, QMIX, and MAVEN fail to learn in our setting.
None of these incorporate novelty-seeking exploration, which are crucial in sparse reward domains.

\vspace{-0.13 in}
\section{Conclusion}
\label{conclusion}
%!TEX root = ms.tex
We propose a framework for designing multi-agent intrinsic reward functions with diverse properties, and compare an varied set on several multi-agent exploration tasks in a gridworld domain as well as in VizDoom.
Overall, we can see that cooperative multi-agent tasks can, in many cases, benefit from intrinsic rewards that take into account what other agents have explored, but there are various ways to incorporate that information, each resulting in different coordinated behaviors.
We show that our method is capable of matching or surpassing the performance of the non-adaptive oracle on various tasks while using the same number of samples collected from the environment.
Furthermore, we show that adaptation of exploration type over the course of training can overcome the limitations of choosing a fixed exploration type.

% \section*{Acknowledgements}

\bibliographystyle{named}
\bibliography{citations}

\appendix
%!TEX root = ms.tex

\section{Environment Details}

\subsection{Gridworld}
At each step there is a 10\% chance of an agent's action being replaced by a random one.
Inspired by the enemies in Montezuma's Revenge which increase the challenge of exploration, we introduce ``wormholes'' which have a probability of opening at each time step.
This probability, which agents can observe, changes at each step using a biased random walk such that it moves toward one, until the hole opens and it resets to zero.
If an agent steps into an open wormhole, they will be sent back to their starting position.
These are an important aspect of the environment, as they add difficulty to exploration.
The probability, $\rho$, of a wormhole opening at each step, $t$, evolves as such: $\rho_{t+1} = \rho_t + \mathcal{N}(\mu, \sigma)$, where $\mu = \sigma = 0.05$ for \textbf{\textsc{Task 1}} and $\mu = \sigma = 0.005$ for \textbf{2} and \textbf{3}.

Agents observe their global position in $(x,y)$ coordinates (scalars), as well as local information regarding walls in adjacent spaces, the probability of their adjacent spaces opening into a wormhole, the relative position of other agents (if they are within 3 spaces), as well as information about which treasures the agent has already collected in the given episode.
The global state is represented by the $(x,y)$ coordinates of all agents, as one-hot encoded vectors for $x$ and $y$ separately, as well as the local information of all agents regarding wormholes, walls, and treasures collected.
Each agent's action space consists of the 4 cardinal directions as well as an option to not move, which is helpful in cases where an agent is waiting for a wormhole to be safe to cross.

\subsection{VizDoom}
We use the ``My Way Home'' map, which has been used as a test bed for single-agent exploration techniques~\citep{pmlr-v70-pathak17a}, and modify it for multi-agent tasks.
Since the agents are moved to a central location closer to their rewards than in the original map, we lower the frame skip from 4 to 2, in order to force agents to take twice as many steps in order to explore the same areas, maintaining the challenging nature of exploration in the original task.
Agents receive their egocentric view in the form of 48x48 grayscale images as observations along with an indicator of which agents (if any) have collected each reward, and we use a vector based global state which includes all agents' $(x,y)$ positions and velocities, their orientations, as well as the same indicator of which agent has collected each reward.
The policies \emph{do not} receive the global state as input, as it is only used for the centralized critics.
As in the gridworld setting, we use count-based intrinsic rewards for VizDoom; however, since VizDoom is not a discrete domain, we separate agents' $(x,y)$ positions into discrete bins and use the counts for these bins.
There are 30 bins in the $x$ dimension and 26 in the $y$ dimension.
$(x,y)$ positions in the global state are represented both as scalars and one-hot vectors indicating which bin the agents are currently occupying.
Each agent can choose from 3 actions at each time step: turn left, turn right, or go forward.

\section{Training Details}
\begin{algorithm}[H]
\caption{Training Procedure for Multi-Explore w/ Soft Actor-Critic~\citep{haarnoja2018soft}}\label{training_alg}
\begin{algorithmic}[1]
    % \begin{varwidth}[t]{\linewidth}
    \STATE Initialize environment with $n$ agents
    \STATE Initialize replay buffer, $D$
    \STATE $t_{\text{update}} \gets 0$
    \STATE $t_{\text{ep}} \gets \text{max ep length}$
    \FOR{$t = 1 \ldots \text{total steps}$}
        \IF{\text{episode done} or $t_{\text{ep}} == \text{max ep length}$}
            \FOR{$j = 1 \ldots \text{niters-selector}$}
                \STATE \textit{UpdateSelector}(R, h) \COMMENT{Eqs 8-9 in main text}
            \ENDFOR
            \STATE $s, \mathbf{o} \gets$ \textit{ResetEnv}()
            \STATE $h \sim \Pi$  \COMMENT{Sample policy head}
            \STATE $t_{\text{ep}} \gets 0$
            \STATE $R \gets 0$
        \ENDIF
        \STATE Select actions $a_i \sim \pi_i^h(\cdot|o_i)$ for each agent, $i$
        \STATE Send actions to environment and get $s'$, $\mathbf{o}'$, $r$
        \STATE $R \gets R + \gamma^{t_\text{ep}}~r$
        \STATE Store transition $(s, \mathbf{o}, \mathbf{a}, s', \mathbf{o}', r)$ in $D$
        \STATE $t_{\text{update}} += 1$
        \STATE $t_{\text{ep}} += 1$
        \IF{$t_{\text{update}} == \text{steps per update}$}
            \FOR{$j = 1 \ldots \text{niters-model}$}
                \STATE Sample minibatch, $B \sim D$
                \STATE \textit{UpdateCritic}($B$) \COMMENT{Eqs 2,4 in main text}
                \STATE \textit{UpdatePolicies}($B$) \COMMENT{Eqs 5-7 in main text}
                \STATE Update target parameters:
                    $$\bar{\Psi} = \tau \bar{\Psi} + (1 - \tau)\Psi$$
                    $$\bar{\Theta} = \tau \bar{\Theta} + (1 - \tau)\Theta$$
            \ENDFOR
            
            \STATE $t_{\text{update}} \gets 0$
        \ENDIF
    \ENDFOR
    % \end{varwidth}
\end{algorithmic}
\end{algorithm}
The training procedure is detailed in Algorithm~\ref{training_alg}, and all hyperparameters are listed in Tables~\ref{tab:hyperparams} and \ref{tab:vzd_hyperparams}.
Hyperparameters were selected by tuning one parameter at a time through intuition on task 1 with 2 agents and then applying to the rest of the settings with minimal changes.
Individual runs used a single NVIDIA Titan Xp GPU for training.
Two machines were used, one (desktop) has an Intel i7-6800K CPU with 128 GB RAM, and another (server) has two Intel Xeon(R) Gold 6154 CPUs with 256 GB RAM.
Both machines run Ubuntu 16.04.
Software versions are provided in the code supplement.

\begin{table*}[h]
\captionsetup{width=0.8\textwidth,justification=centering}
\begin{center}
    % \small
    \caption{Hyperparameter settings across all runs in gridworld.}
    \begin{tabular}{@{}lll@{}}
        \toprule
        Name & Description & Value \\
        \midrule
        $Q$ lr & learning rate for centralized critic & 0.001 \\
        $Q$ optimizer & optimizer for centralized critic & Adam~\citep{kingma2014adam} \\
        $\pi$ lr & learning rate for decentralized policies & 0.001 \\
        $\pi$ optimizer & optimizer for decentralized policies & Adam \\
        $\Pi$ lr & learning rate for policy selector & 0.04 \\
        $\Pi$ optimizer & optimizer for policy selector & SGD \\
        $\tau$ & target function update rate & 0.005 \\
        bs & batch size & 1024 \\
        \midrule
        total steps & number of total environment steps & $1e6$ \\
        steps per update & number of environment steps between updates & 100 \\
        niters-model & number of iterations per update for policies and critics & 50 \\
        niters-selector & number of iterations per update for policy selector & 50$^a$/2$^b$ \\
        max ep length & maximum length of an episode before resetting & 500 \\
        \midrule
        \multirow{2}{*}{$\Psi$ penalty} & coefficient for weight decay on & \multirow{2}{*}{0.001} \\
                                        & parameters of Q-function                               & \\
        \multirow{2}{*}{$\Theta$ penalty} & coefficient on $L_2$ penalty on pre-softmax & \multirow{2}{*}{0.001} \\
                                        & output of policies                               & \\
        \multirow{2}{*}{$\theta$ penalty} & coefficient for weight decay on & \multirow{2}{*}{0.001} \\
                                        & parameters of policy selector                               & \\
        \midrule
        $|D|$ & maximum size of replay buffer & $1e6$ \\
        $\alpha$ & action policy reward scale & 100 \\
        $\eta$ & selector policy reward scale & 5$^a$/0.1$^b$ \\
        $\gamma$ & discount factor & 0.99 \\
        \midrule
        $\beta$ & relative weight of intrisic rewards to extrinsic & 0.1 \\
        $\zeta$ & decay rate of count-based rewards & 0.7 \\
        \bottomrule
    \end{tabular}
    \label{tab:hyperparams}
\end{center}
\begin{center}
$^a$: \textbf{\textsc{Tasks 1-3}}, $^b$: \textbf{\textsc{Flip-Task}}
\end{center}
\end{table*}

\begin{table*}[h]
\captionsetup{width=0.8\textwidth,justification=centering}
\begin{center}
    % \small
    \caption{Hyperparameter settings across all runs in VizDoom (only where different from Table~\ref{tab:hyperparams}).}
    \begin{tabular}{@{}lll@{}}
        \toprule
        Name & Description & Value \\
        \midrule
        $Q$ lr & learning rate for centralized critic & 0.0005 \\
        $\pi$ lr & learning rate for decentralized policies & 0.0005 \\
        bs & batch size & 128 \\
        \midrule
        $|D|$ & maximum size of replay buffer & $5e5$ \\
        \bottomrule
    \end{tabular}
    \label{tab:vzd_hyperparams}
\end{center}
\end{table*}

\section{Network Architectures}
\label{network_arch}
In this section we list, in pseudo-code, the architectures we used for all policies and critics
\subsection{Gridworld}
$\theta_i^\text{share}$ (shared for policy heads):
\begin{lstlisting}
obs_size = observations.shape[1]
fc1 = Linear(in_dim=obs_size, out_dim=128)
nl1 = ReLU()
\end{lstlisting}

$\theta_i^j$ (specific to each policy head):
\begin{lstlisting}
n_acs = actions.shape[1]
fc2 = Linear(in_dim=fc1.out_dim, out_dim=32)
nl2 = ReLU()
fc3 = Linear(in_dim=fc2.out_sim, out_dim=n_acs)
\end{lstlisting}

$\psi^\text{share}$ (shared across critics for all agents and reward types):
\begin{lstlisting}
state_size = states.shape[1]
fc1 = Linear(in_dim=state_size, out_dim=128)
nl1 = ReLU()
\end{lstlisting}

$\psi_{i,j}$ (specific to each agent/policy head combination, same architecture for extrinsic and intrinsic critics):
\begin{lstlisting}
n_acs = actions.shape[1]
# fc2 takes other agents' actions as input
fc2 = Linear(in_dim=fc1.out_dim + (num_agents - 1) * n_acs, out_dim=128)
nl2 = ReLU()
fc3 = Linear(in_dim=fc2.out_dim, out_dim=n_acs)
\end{lstlisting}

\subsection{VizDoom}
$\theta_i^\text{share}$ (shared for policy heads belonging to one agent):
\begin{lstlisting}
# vector observation encoder
vect_obs_size = vector_observations.shape[1]
vect_fc = Linear(in_dim=obs_size, out_dim=32)
vect_nl = ReLU()
# image observation encoder
img_obs_channels = image_observations.shape[1]
pad1 = ReflectionPadding(size=1)
conv1 = Conv2D(in_channels=img_obs_channels, out_channels=32, filter_size=3, stride=2)
conv_nl1 = ReLU()
pad2 = ReflectionPadding(size=1)
conv2 = Conv2D(in_channels=conv1.out_channels, out_channels=32, filter_size=3, stride=2)
conv_nl2 = ReLU()
pad3 = ReflectionPadding(size=1)
conv3 = Conv2D(in_channels=conv2.out_channels, out_channels=32, filter_size=3, stride=2)
conv_nl3 = ReLU()
pad4 = ReflectionPadding(size=1)
conv4 = Conv2D(in_channels=conv3.out_channels, out_channels=32, filter_size=3, stride=2)
conv_nl4 = ReLU()
conv_flatten = Flatten()  # flatten output of conv layers
conv_fc = Linear(in_dim=conv_flatten.out_dim, out_dim=128)
conv_fc_nl = ReLU()
\end{lstlisting}

$\theta_i^j$ (specific to each policy head):
\begin{lstlisting}
n_acs = actions.shape[1]
# takes concatenation of image and vector encodings as input
fc_out1 = Linear(in_dim=conv_fc.out_dim + vect_fc.out_dim, out_dim=32)
fc_out_nl = ReLU()
fc_out2 = Linear(in_dim=fc_out1.out_dim, out_dim=n_acs)
\end{lstlisting}

$\psi^\text{share}$ (shared across critics for all agents and reward types):
\begin{lstlisting}
state_size = states.shape[1]
fc1 = Linear(in_dim=state_size, out_dim=256)
nl1 = ReLU()
\end{lstlisting}

$\psi_{i,j}$ (specific to each agent/policy head combination, same architecture for extrinsic and intrinsic critics):
\begin{lstlisting}
n_acs = actions.shape[1]
# fc2 takes other agents' actions as input
fc2 = Linear(in_dim=fc1.out_dim + (num_agents - 1) * n_acs, out_dim=256)
nl2 = ReLU()
fc3 = Linear(in_dim=fc2.out_dim, out_dim=n_acs)
\end{lstlisting}

\section{Training Curves}
\label{training_curves}

\begin{table*}[t]
\captionsetup{width=0.8\textwidth,justification=centering}
\begin{center}
    \small
    \caption{\# of treasures found  with standard deviation across 6 runs. Scores are bolded where the best mean score falls within one standard deviation. Our method (\textbf{\textsc{multi}}) matches or outperforms the non-adaptive oracle in nearly all settings}
    \begin{tabular}{@{}llllllll@{}}
        \toprule
        \multicolumn{8}{c}{\textbf{\textsc{Gridworld}}} \\
        \midrule
        && \multicolumn{6}{c}{Intrinsic reward type} \\
        \cmidrule(l){3-8} Task & $n$ & \textsc{Independent} & \textsc{Minimum} & \textsc{Covering} & \textsc{Burrowing} & \textsc{Lead-Follow} & \textbf{\textsc{Multi}} \\
        \midrule
        1 & 2 & 0.14 $\pm$ 0.05 & \textbf{1.62 $\pm$ 0.59} & 0.13 $\pm$ 0.12 & \textbf{1.98 $\pm$ 0.06} & 0.18 $\pm$ 0.24 & \textbf{2.00 $\pm$ 0.00} \\
          & 3 & 1.16 $\pm$ 0.11 & \textbf{1.49 $\pm$ 0.76} & 0.00 $\pm$ 0.00 & \textbf{2.06 $\pm$ 1.05} & 0.34 $\pm$ 0.45 & \textbf{2.23 $\pm$ 0.73} \\
          & 4 & 0.84 $\pm$ 0.29 & \textbf{1.78 $\pm$ 0.44} & 0.00 $\pm$ 0.00 & \textbf{1.90 $\pm$ 0.49} & 1.17 $\pm$ 0.39 & \textbf{2.04 $\pm$ 0.61 }\\
        \midrule
        2 & 2 & \textbf{2.00 $\pm$ 0.00} & 0.92 $\pm$ 0.10 & \textbf{1.11 $\pm$ 0.99} & 0.98 $\pm$ 0.05 & \textbf{1.73 $\pm$ 0.66} & \textbf{1.83 $\pm$ 0.41} \\
          & 3 & \textbf{2.66 $\pm$ 0.80} & 1.11 $\pm$ 0.29 & 0.54 $\pm$ 0.80 & 1.80 $\pm$ 0.29 & \textbf{3.00 $\pm$ 0.00} & 1.80 $\pm$ 0.71 \\
          & 4 & \textbf{1.83 $\pm$ 1.08} & 0.93 $\pm$ 0.13 & 0.22 $\pm$ 0.18 & \textbf{1.99 $\pm$ 0.67} & \textbf{2.66 $\pm$ 2.06} & \textbf{2.54 $\pm$ 1.21} \\
        \midrule
        3 & 2 & \textbf{1.39 $\pm$ 0.94} & {0.67 $\pm$ 1.03} & 0.29 $\pm$ 0.37 & {0.67 $\pm$ 1.03} & 0.83 $\pm$ 0.67 & \textbf{2.00 $\pm$ 0.00} \\
          & 3 & \textbf{1.68 $\pm$ 0.70} & 0.60 $\pm$ 0.73 & 0.09 $\pm$ 0.08 & \textbf{1.35 $\pm$ 1.16} & \textbf{1.59 $\pm$ 0.83} & \textbf{2.21 $\pm$ 0.91} \\
          & 4 & 1.12 $\pm$ 0.47 & 1.36 $\pm$ 0.71 & 0.05 $\pm$ 0.05 & \textbf{2.14 $\pm$ 1.49} & 0.68 $\pm$ 0.53 & \textbf{1.73 $\pm$ 0.47} \\
        \midrule
        Flip-Task & 2 & \textbf{3.04 $\pm$ 1.33} & \textbf{2.66 $\pm$ 1.51} & 2.16 $\pm$ 1.47 & 1.96 $\pm$ 1.07 & 2.73 $\pm$ 1.26 & \textbf{4.03 $\pm$ 0.97} \\
        \midrule
        \multicolumn{8}{c}{\textbf{\textsc{VizDoom}}} \\
        \midrule
        1 & 2 & 0.94 $\pm$ 0.54 & \textbf{1.57 $\pm$ 0.74} & 0.16 $\pm$ 0.17 & \textbf{1.94 $\pm$ 0.10} & 0.61 $\pm$ 0.43 & \textbf{1.98 $\pm$ 0.03} \\
        2 &   & \textbf{1.52 $\pm$ 0.75} & \textbf{1.53 $\pm$ 0.74} & 0.70 $\pm$ 1.00 & {0.63 $\pm$ 0.04} & \textbf{1.93 $\pm$ 0.10} & {1.23 $\pm$ 0.65} \\
        3 &   & 0.18 $\pm$ 0.19 & \textbf{0.64 $\pm$ 1.05} & 0.45 $\pm$ 0.46 & {0.29 $\pm$ 0.25} & 0.20 $\pm$ 0.17 & \textbf{1.64 $\pm$ 0.63 }\\
        \bottomrule
    \end{tabular}
    \label{tab:all_task_res_full}
\end{center}
\end{table*}

\subsection{Gridworld}

\begin{figure}[H]
    \centering
    \includegraphics[width=\linewidth]{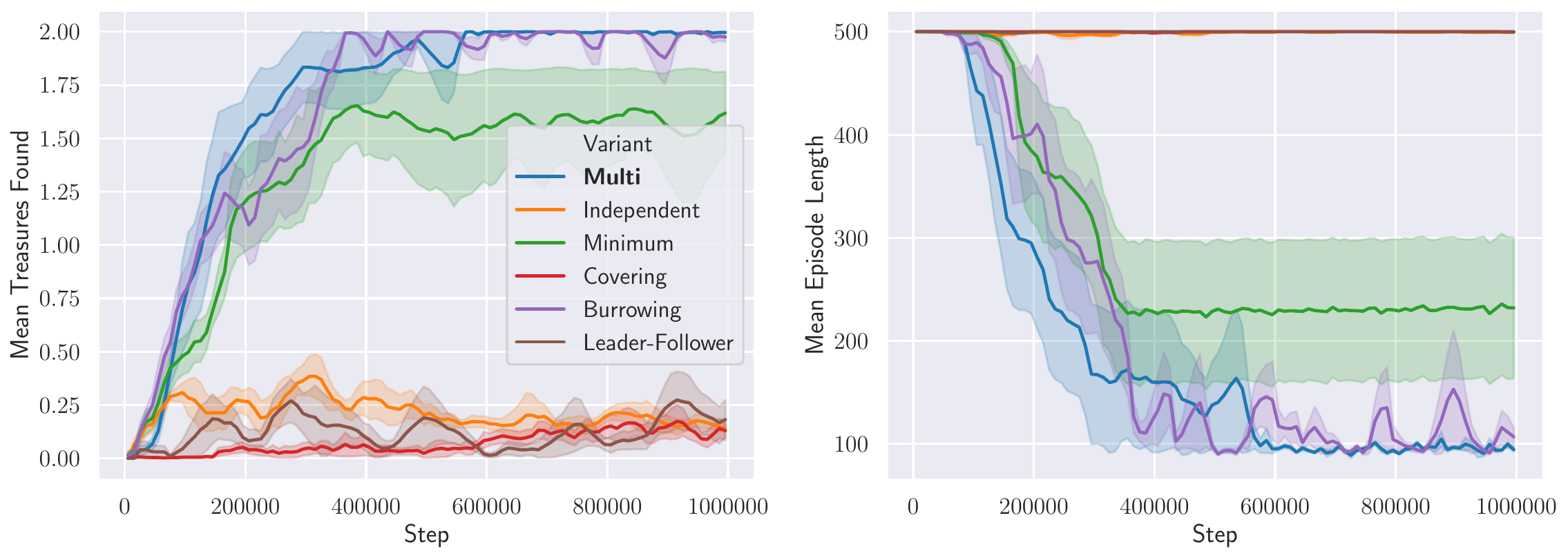}
    \caption{\small Results on Task 1 in Gridworld with 2 agents.}
    \label{fig:15_2_1_res}
\end{figure}

\begin{figure}[H]
    \centering
    \includegraphics[width=\linewidth]{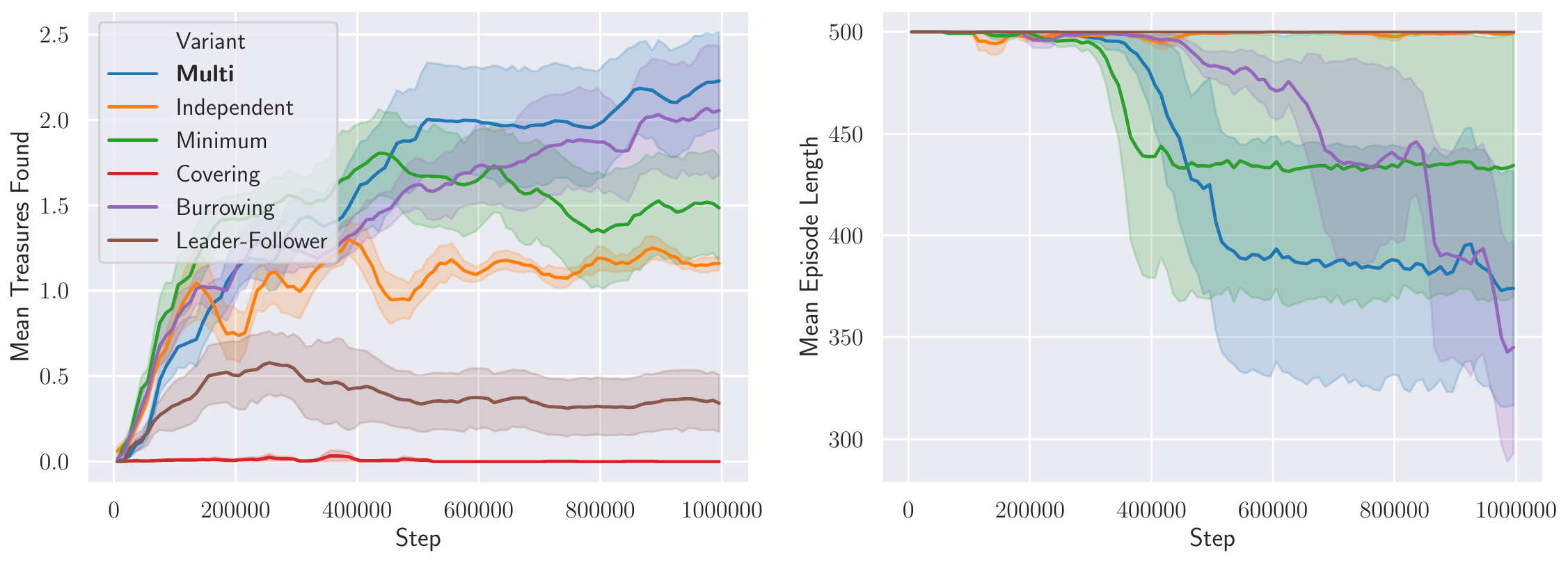}
    \caption{\small Results on Task 1 in Gridworld with 3 agents.}
    \label{fig:15_3_1_res}
\end{figure}

\begin{figure}[H]
    \centering
    \includegraphics[width=\linewidth]{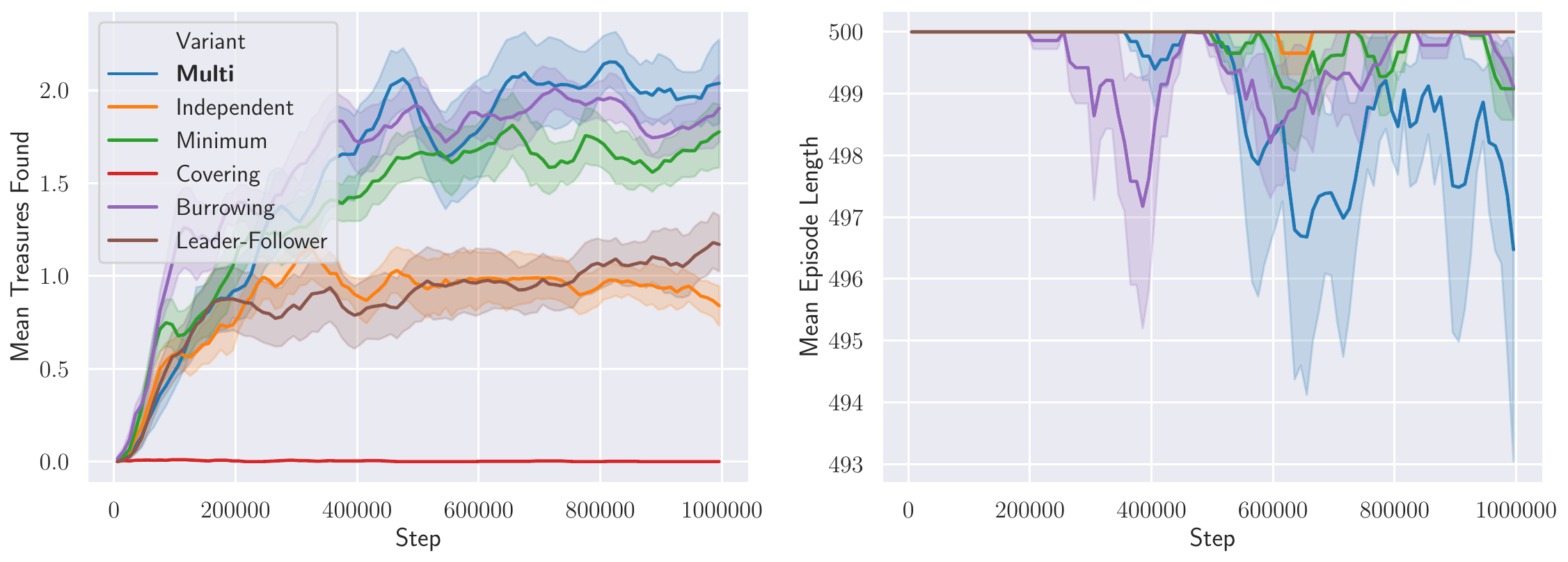}
    \caption{\small Results on Task 1 in Gridworld with 4 agents.}
    \label{fig:15_4_1_res}
\end{figure}

\begin{figure}[H]
    \centering
    \includegraphics[width=\linewidth]{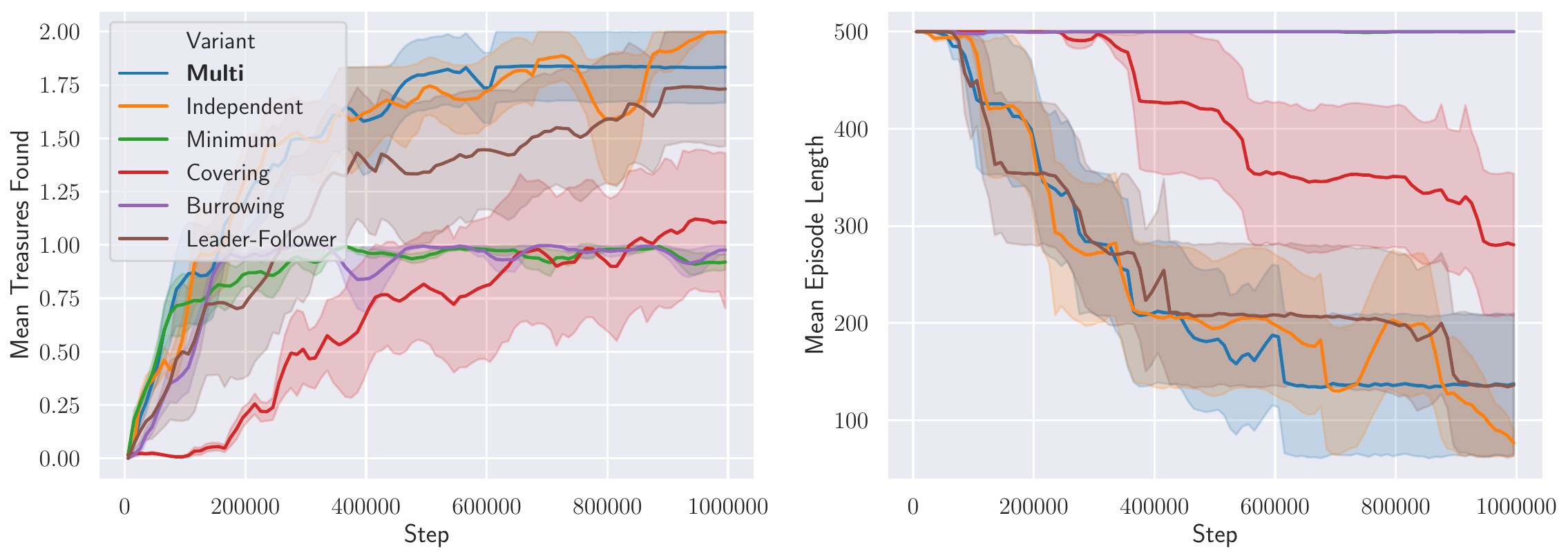}
    \caption{\small Results on Task 2 in Gridworld with 2 agents.}
    \label{fig:15_2_2_res}
\end{figure}

\begin{figure}[H]
    \centering
    \includegraphics[width=\linewidth]{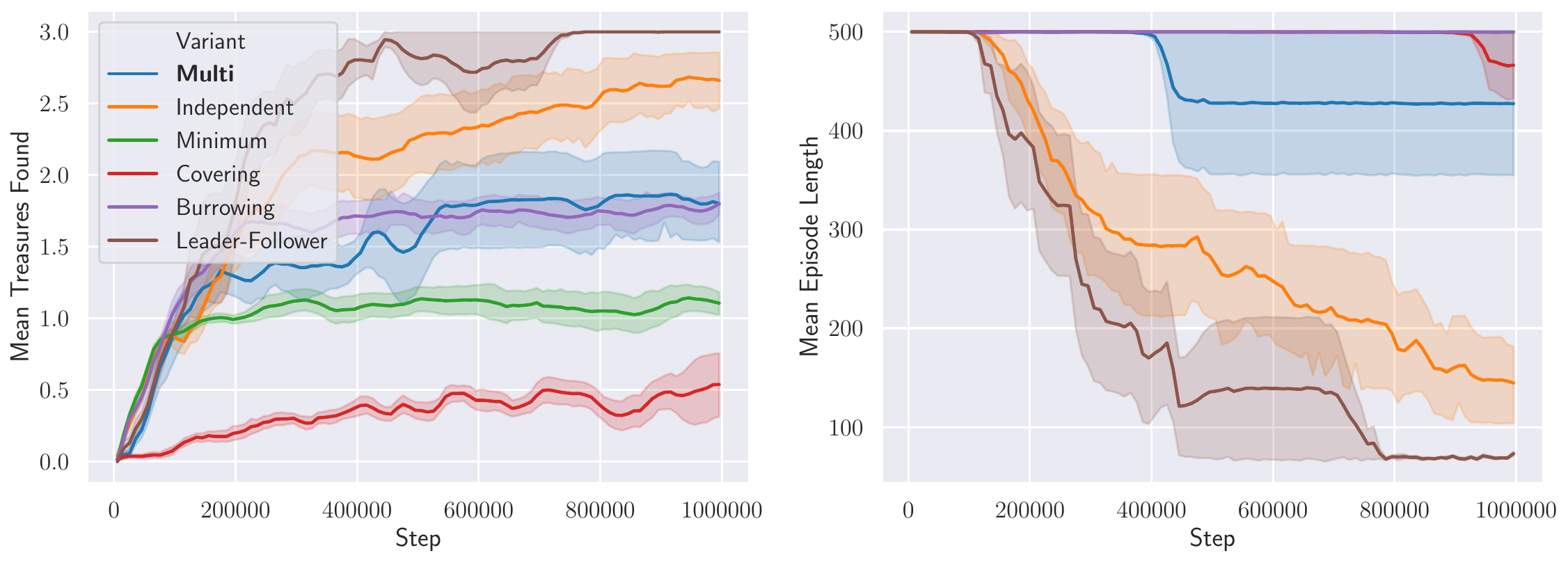}
    \caption{\small Results on Task 2 in Gridworld with 3 agents.}
    \label{fig:15_3_2_res}
\end{figure}

\begin{figure}[H]
    \centering
    \includegraphics[width=\linewidth]{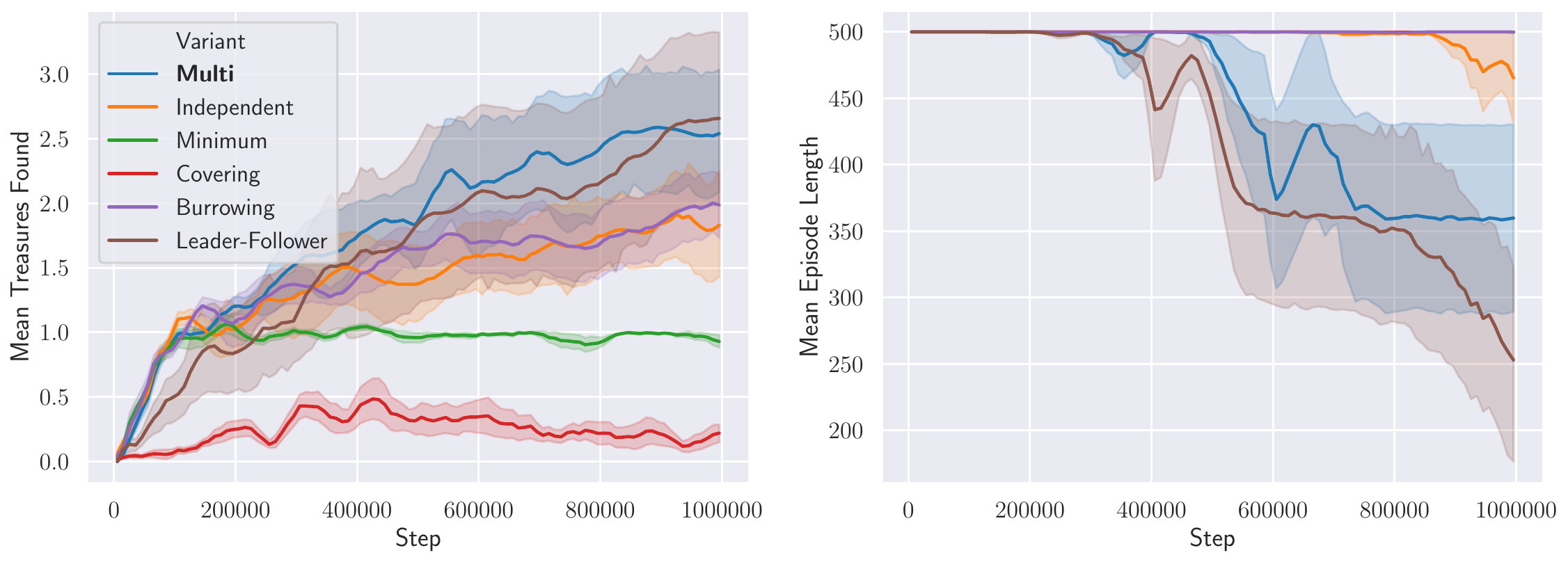}
    \caption{\small Results on Task 2 in Gridworld with 4 agents.}
    \label{fig:15_4_2_res}
\end{figure}

\begin{figure}[H]
    \centering
    \includegraphics[width=\linewidth]{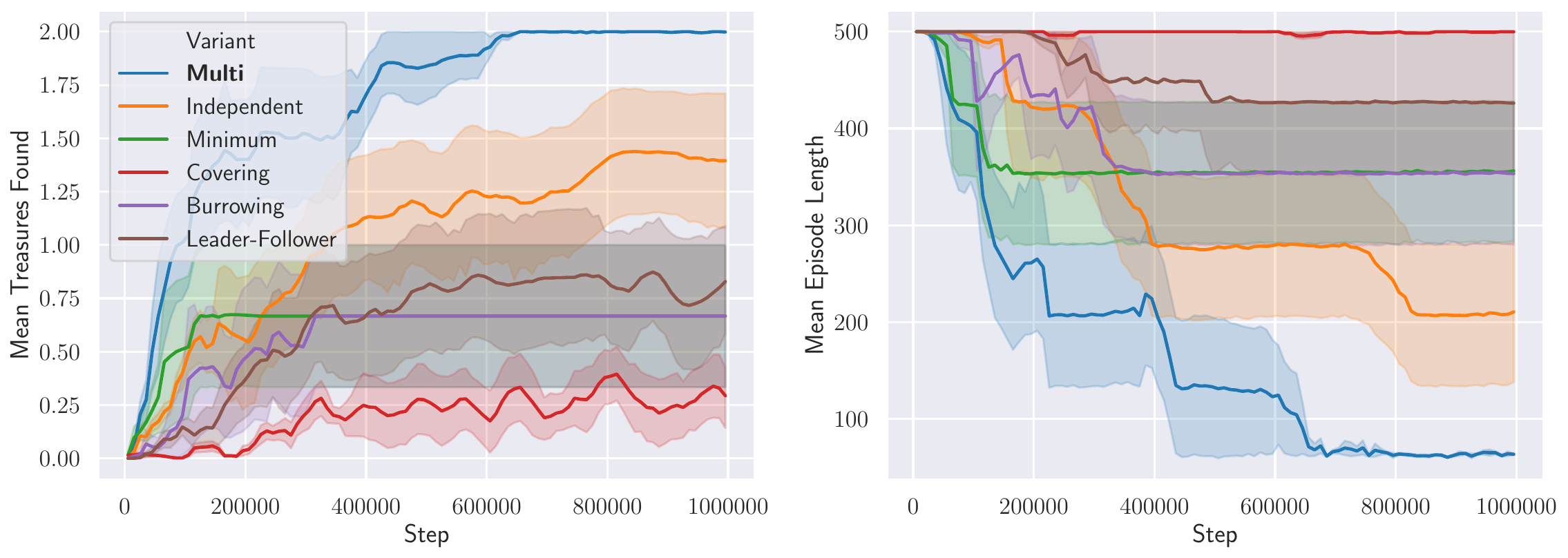}
    \caption{\small Results on Task 3 in Gridworld with 2 agents.}
    \label{fig:15_2_3_res}
\end{figure}

\begin{figure}[H]
    \centering
    \includegraphics[width=\linewidth]{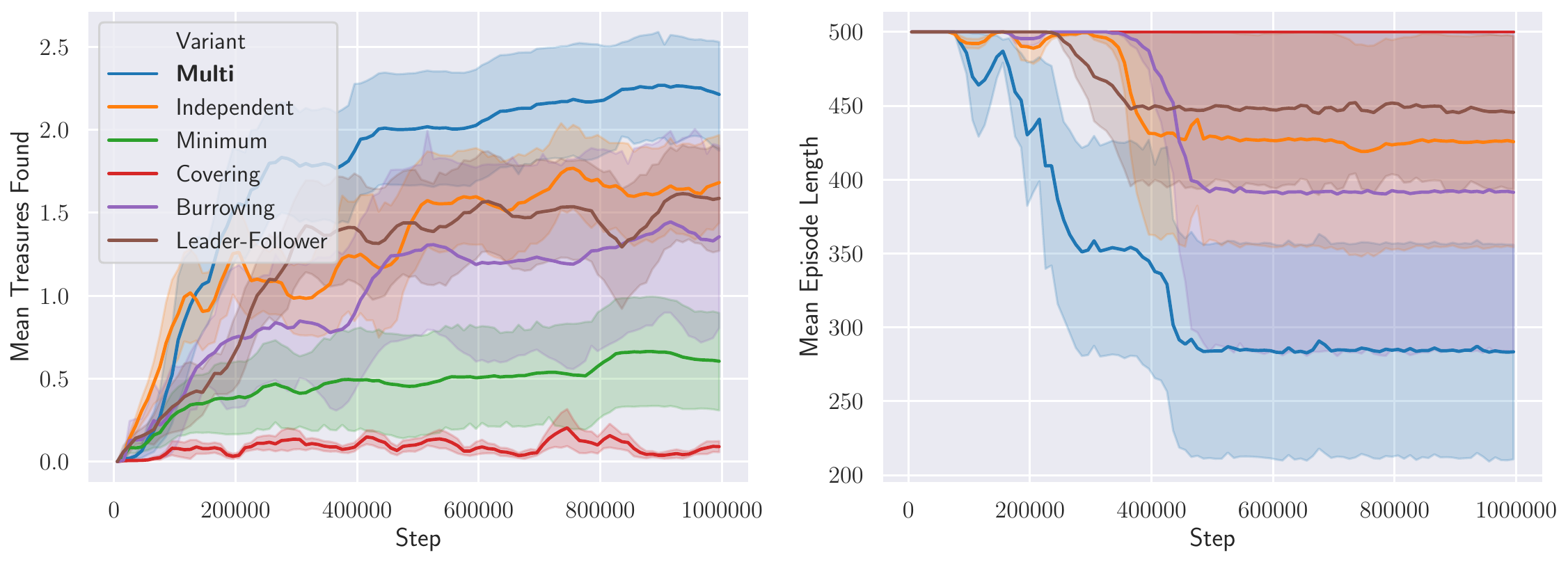}
    \caption{\small Results on Task 3 in Gridworld with 3 agents.}
    \label{fig:15_3_3_res}
\end{figure}

\begin{figure}[H]
    \centering
    \includegraphics[width=\linewidth]{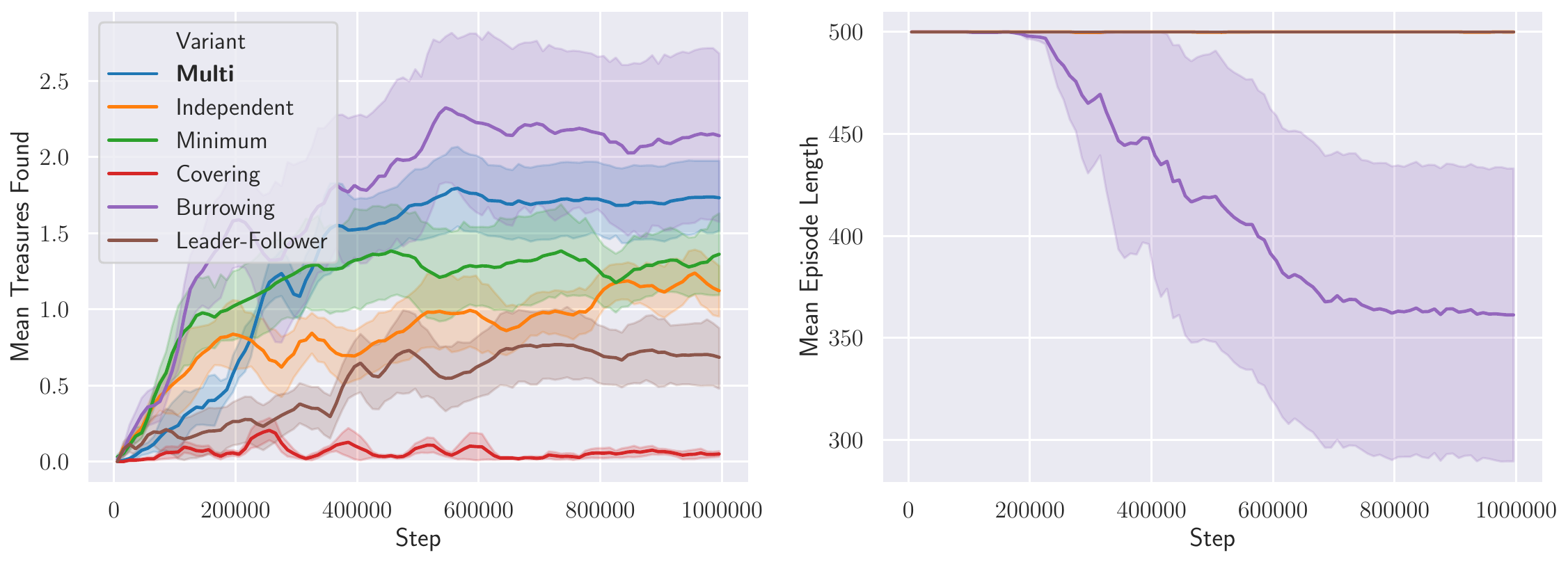}
    \caption{\small Results on Task 3 in Gridworld with 4 agents.}
    \label{fig:15_4_3_res}
\end{figure}

\subsection{Vizdoom}

\begin{figure}[H]
    \centering
    \includegraphics[width=\linewidth]{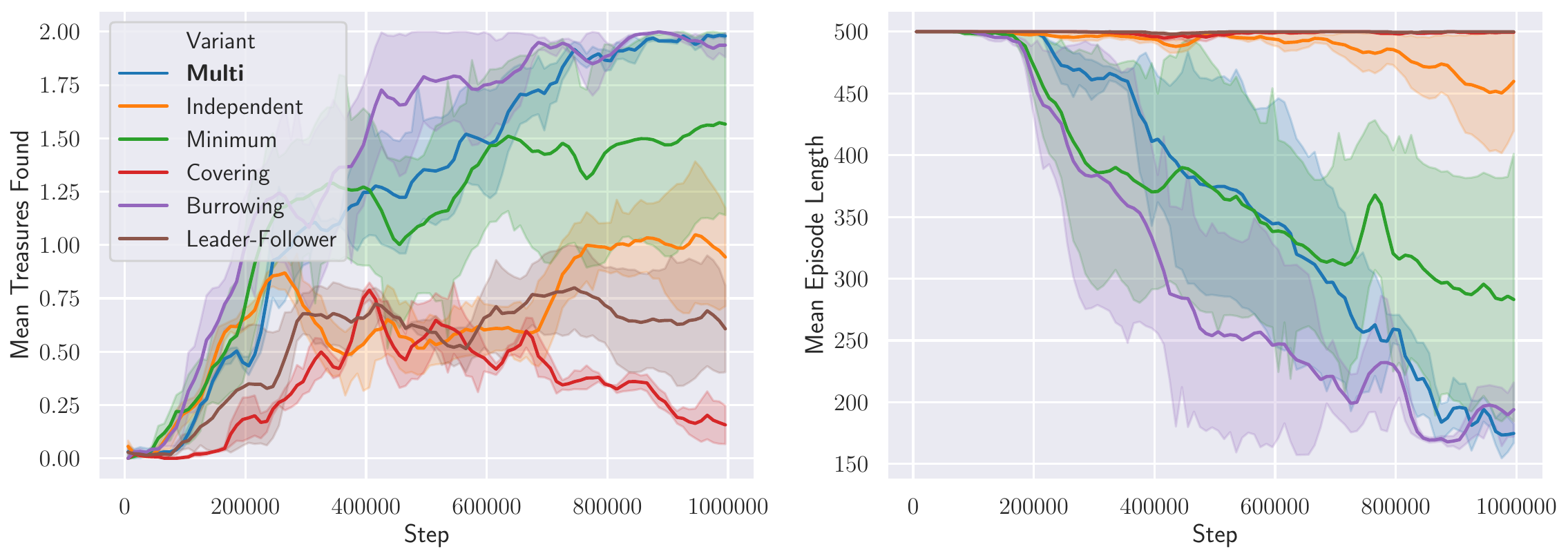}
    \caption{\small Results on Task 1 in Vizdoom.}
    \label{fig:vzd_1_res}
\end{figure}

\begin{figure}[H]
    \centering
    \includegraphics[width=\linewidth]{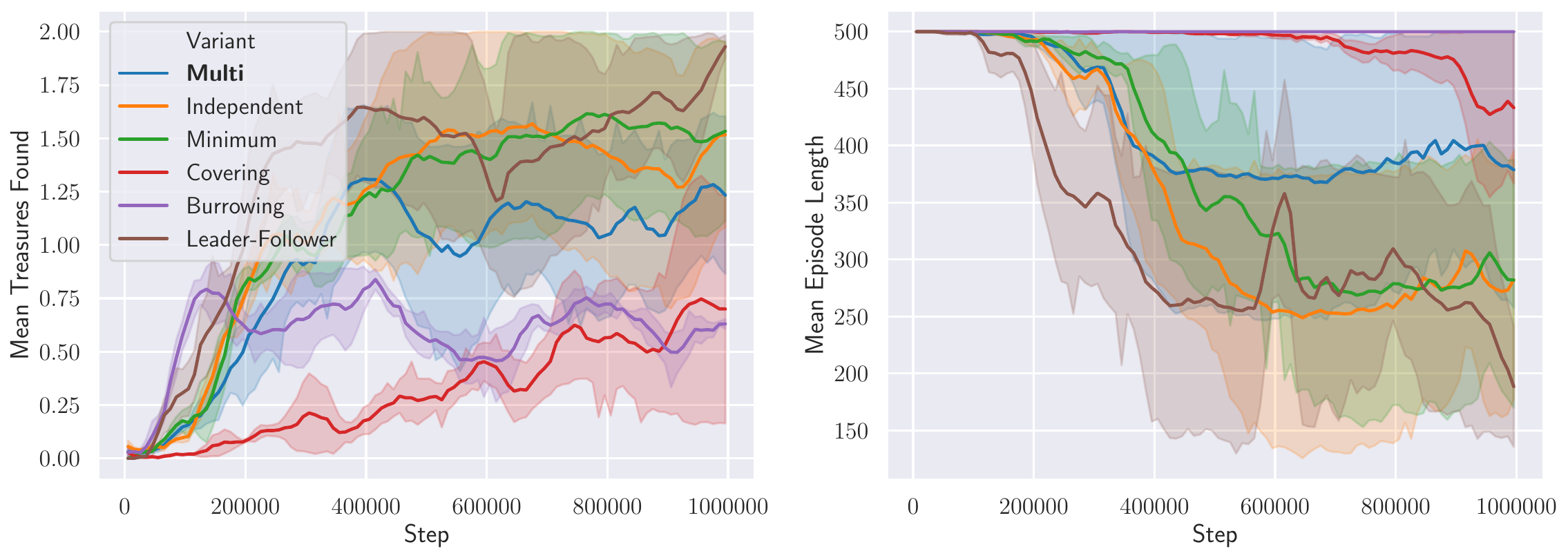}
    \caption{\small Results on Task 2 in Vizdoom.}
    \label{fig:vzd_2_res}
\end{figure}

\begin{figure}[H]
    \centering
    \includegraphics[width=\linewidth]{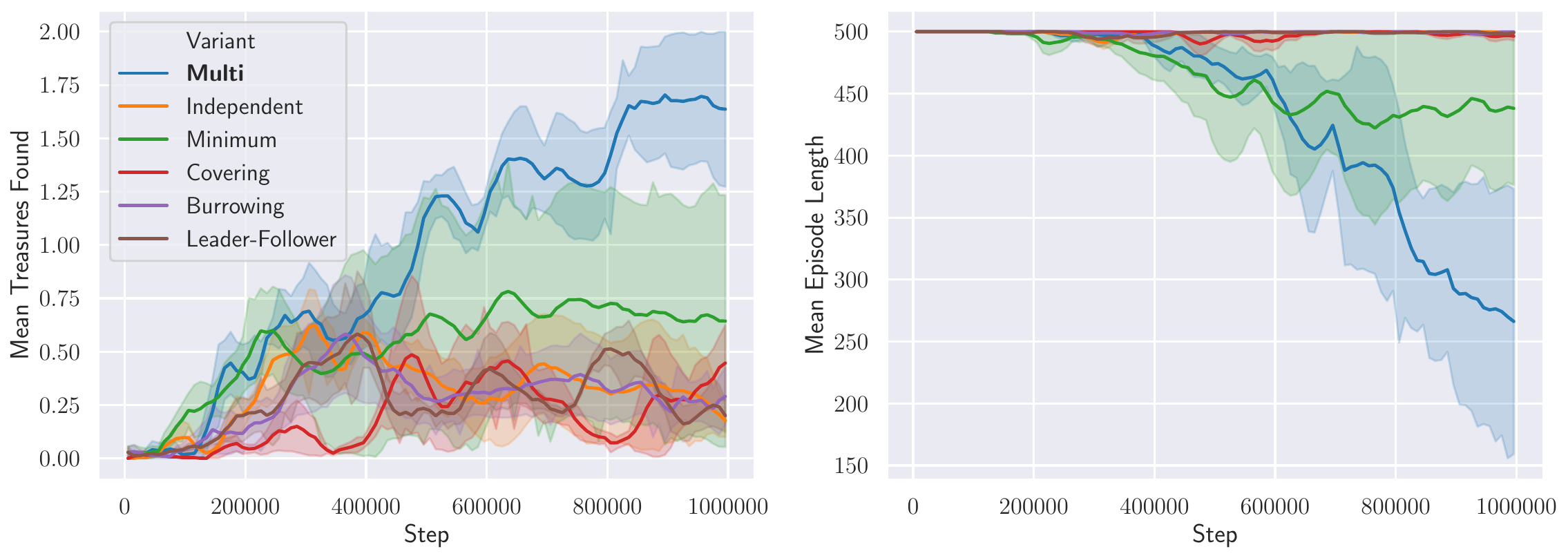}
    \caption{\small Results on Task 3 in Vizdoom.}
    \label{fig:vzd_3_res}
\end{figure}

\section{Discussion of Multi-Agent Intrinsic Rewards and Their Behaviors}
We found the reward types of the non-adaptive oracles (i.e. the reward type that performs best when used exclusively as an intrinsic reward) to largely align with the intuitions behind the design of the tasks.
\textsc{independent} rewards, as expected, result in agents exploring the whole state space without taking other agents into consideration.
Interestingly, policies trained with these rewards can fail on~\textbf{\textsc{Task 3}}, which assigns independent goals to each agent and does not require coordination.
Since our maps contain branching non-linear paths, we find that these agents occasionally suffer from detachment~\citep{ecoffet2019go} (ie. forgetting how to return to an previously explored area).
Both \textsc{minimum} and \textsc{burrowing} encourage agents to not explore the same regions redundantly and are thus successful on \textbf{\textsc{Task 1}} which requires divide-and-conquer type behavior.
\textsc{Leader-Follower} encourages the opposite behavior and is thus most successful on \textbf{\textsc{Task 2}}.
\textsc{covering} rewards, as expected, lead to behavior where agents are constantly switching the regions they explore.
While this behavior alone does not prove to be useful in the tasks we test since the switching slows down overall exploration progress, it serves as a useful mechanism to break out of bad local optima for our meta-policy approach.
The unique behavior of each reward type is visualized in the attached videos.

\section{More Ablations}
\label{more_ablations}
In this section we consider more ablations/comparisons to our model across all three tasks in the 2 agent version of gridworld.
In the first (\textit{Centralized}) we compute intrinsic rewards under the assumption that all agents are treated as one agent.
In other words, we use the inverse count of the number of times that \textbf{all} agents have jointly taken up their combined positions, rather than considering agents independently.
While this reward function will ensure that the global state space is thoroughly searched, it lacks the inductive biases toward spatial coordination that our reward functions incorporate.
As such, it does not learn as efficiently as our method in any of the three tasks.
In the second (\textit{Multi (No Entropy)}) we remove the entropy term from the head selector loss function in order to test its importance.
We find that this ablation is unable to match the performance of the full method, indicating that entropy is crucial in making sure that our method does not converge early to a suboptimal policy selector.
We also compare to the underlying algorithm without intrinsic rewards, multi-agent soft actor-critic (MA-SAC), QMIX~\citep{rashid2018qmix}, and MAVEN~\citep{mahajan2019maven}.
None of these methods use intrinsic rewards, and we find that they are unable to learn in our setting on any of the tasks, highlighting the significant challenge posed.

\begin{figure}[H]
    \centering
    \includegraphics[width=\linewidth]{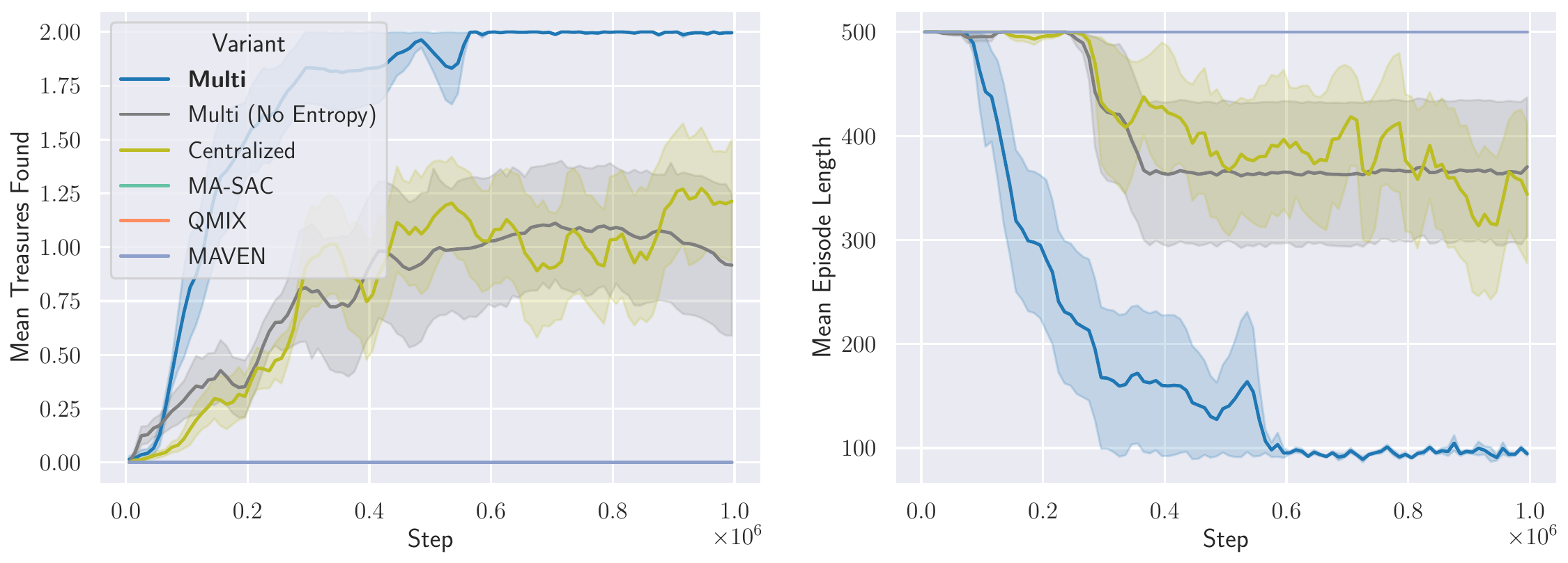}
    \caption{\small Ablations/Baselines on Task 1 in Gridworld with 2 agents.}
    \label{fig:15_2_1_rebuttal}
\end{figure}

\begin{figure}[H]
    \centering
    \includegraphics[width=\linewidth]{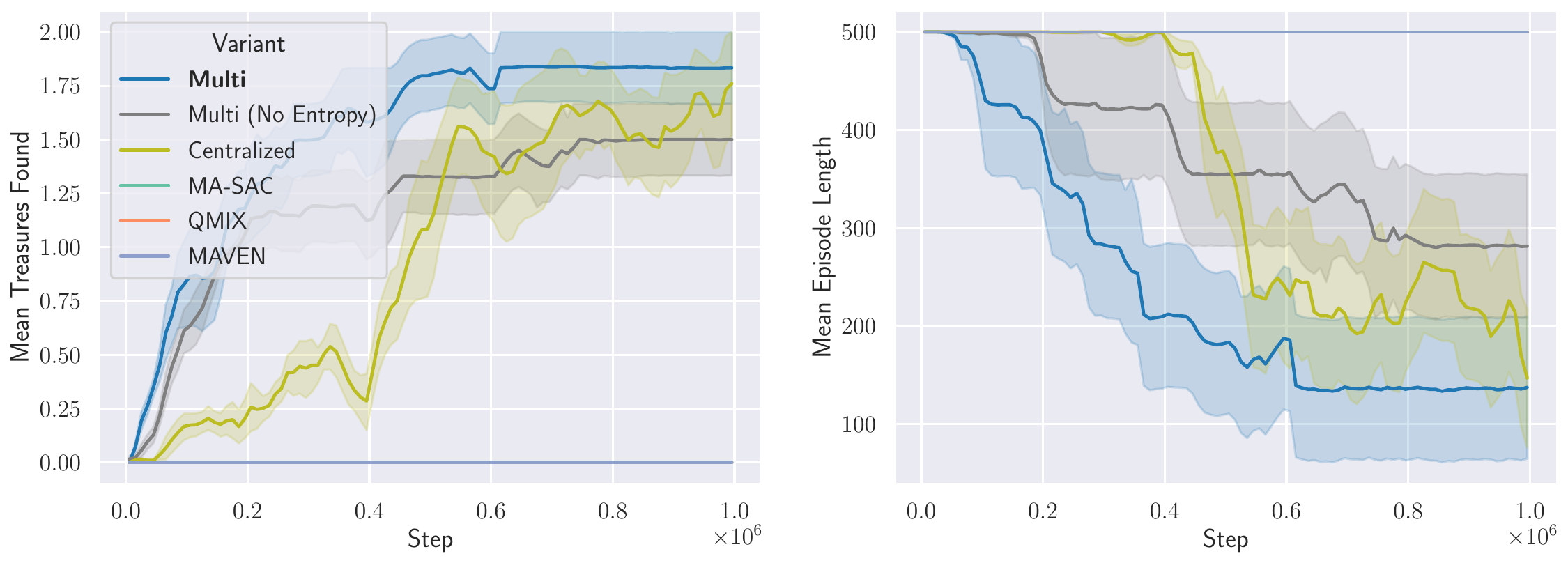}
    \caption{\small Ablations/Baselines on Task 2 in Gridworld with 2 agents.}
    \label{fig:15_2_2_rebuttal}
\end{figure}

\begin{figure}[H]
    \centering
    \includegraphics[width=\linewidth]{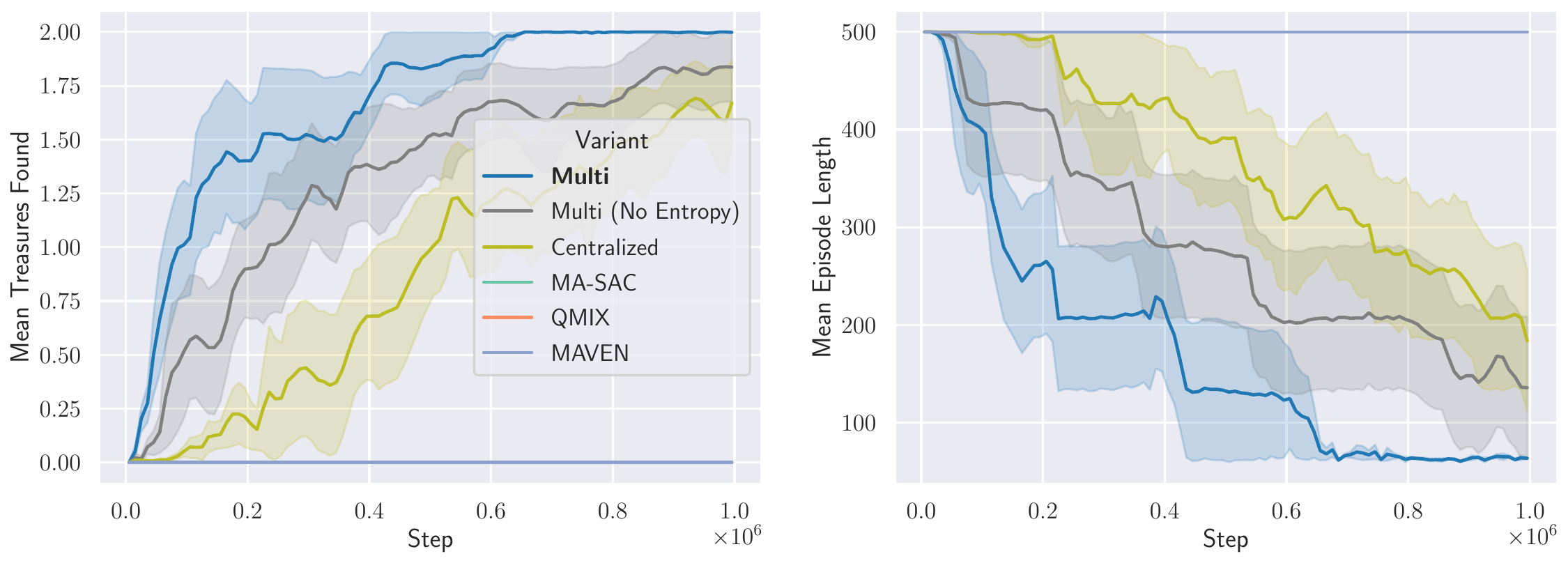}
    \caption{\small Ablations/Baselines on Task 3 in Gridworld with 2 agents.}
    \label{fig:15_2_3_rebuttal}
\end{figure}

\section{Analyzing Meta-Policy}
\label{analyze_metapol}

\begin{figure*}[t]
    \centering
    \begin{subfigure}[t]{0.45\textwidth}
        \centerline{\includegraphics[width=\linewidth]{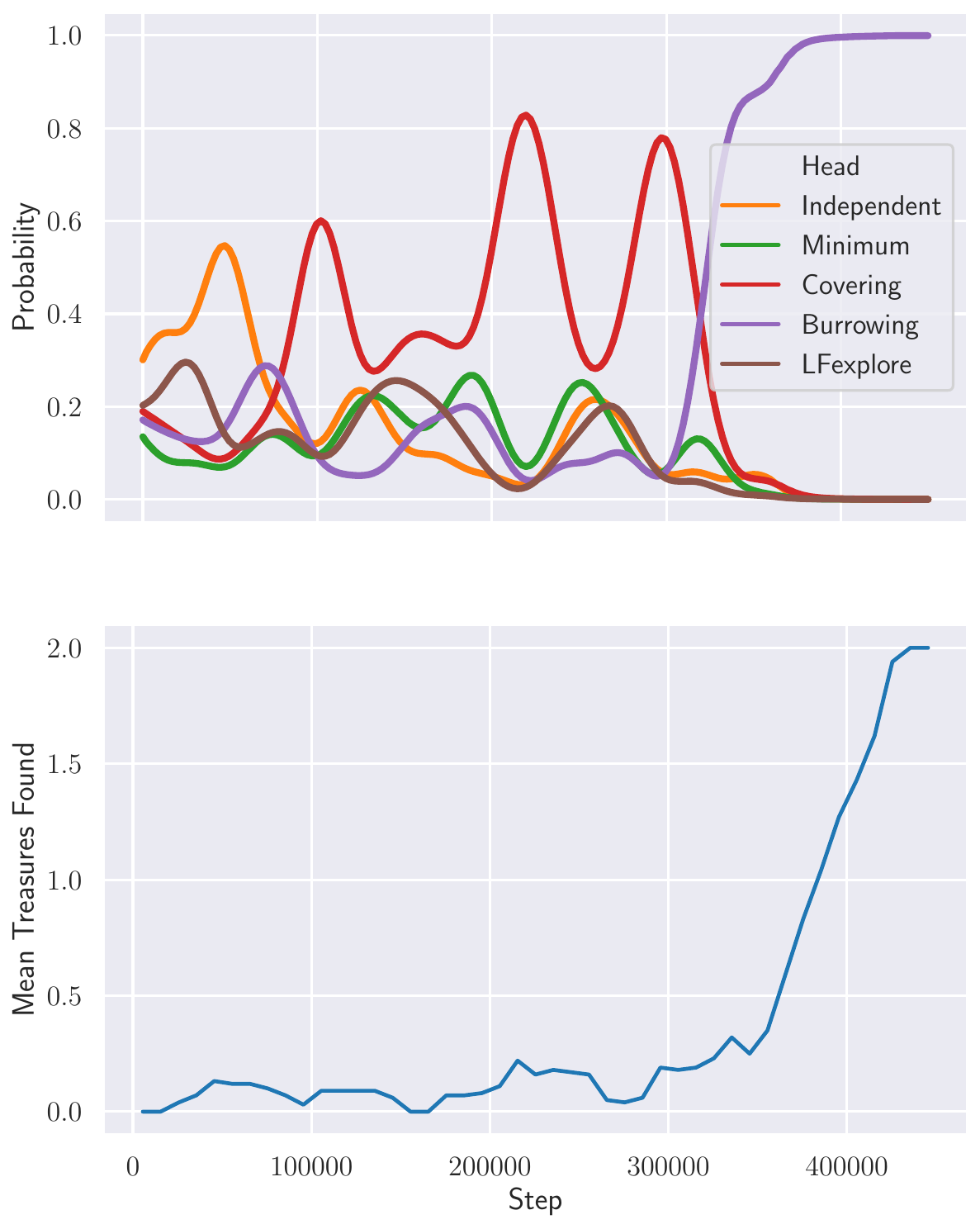}}
        \caption{}
        \label{fig:15_2_3_selector1}
    \end{subfigure}
    \begin{subfigure}[t]{0.45\textwidth}
        \centerline{\includegraphics[width=\linewidth]{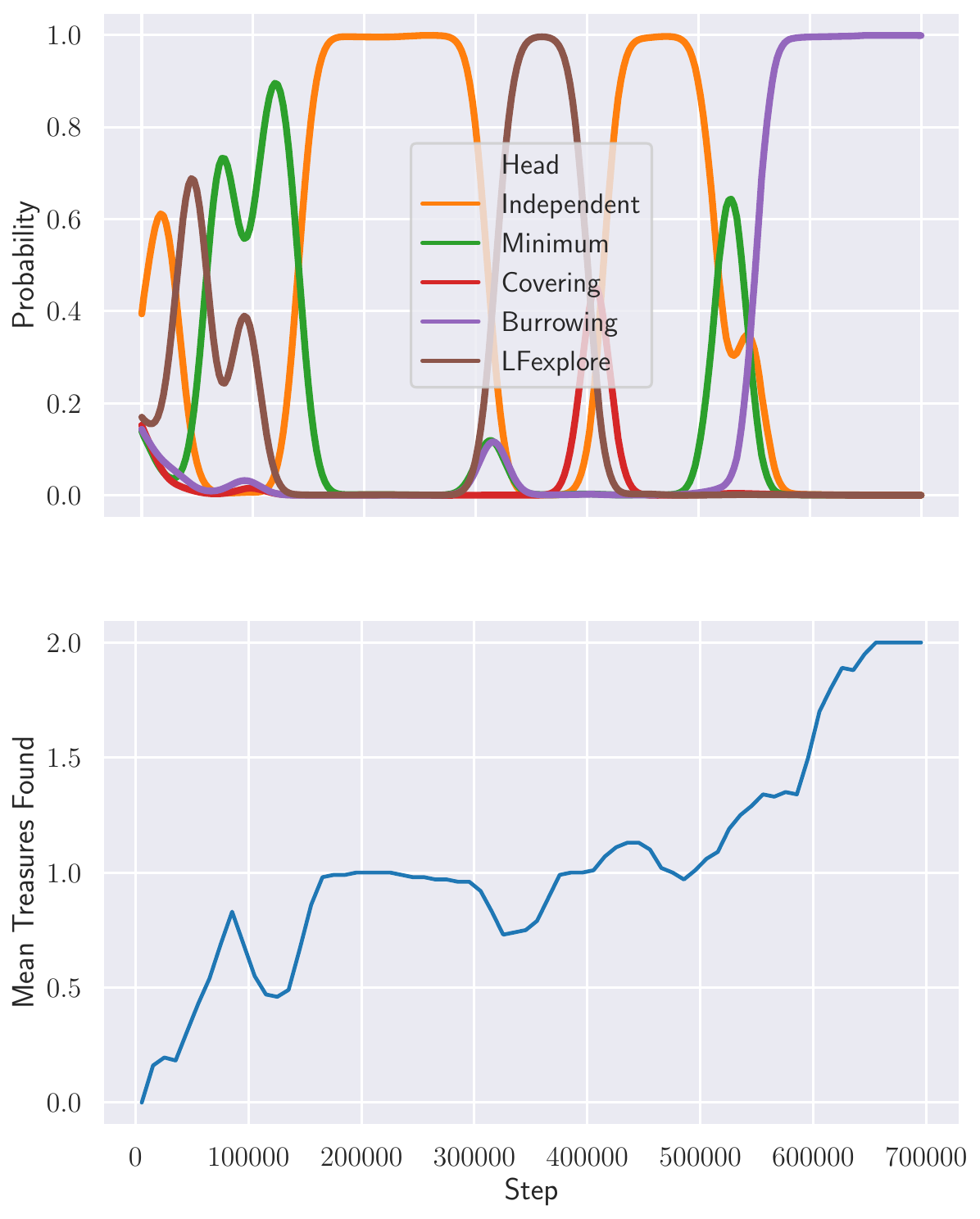}}
        \caption{}
        \label{fig:15_2_3_selector2}
    \end{subfigure}
    \caption{\small Two runs of our method on GridWorld Task 3 with 2 agents.
             Top row shows the evolution of the meta-policy's probability of selecting each policy head.
             Bottom row shows the number of treasures found per episode.
        }
    \label{fig:15_2_3_selector}
\end{figure*}

In Figure~\ref{fig:15_2_3_selector} we analyze the behavior of the meta-policy in two separate runs.
We evaluate on Task 3, since we find that our method is able to surpass the non-adaptive oracle.
This task assigns specific goals to each agent.
As such, one might expect that independent exploration would work most effectively in this setting.
While independent exploration is effective (see Figure~\ref{fig:15_2_3_res}), we find that our method can outperform it.
In both runs, we find that burrowing rewards are selected when the agents finally learn how to solve the task; however, we find that burrowing rewards are not necessarily successful when deployed on their own.
This lack of success is likely due to the fact that these rewards cause the agents to pick a region and commit to exploring it for the duration of training.
As such, the agents may pick the ``wrong'' region at first and never be able to recover.
On the other hand, using our methods, the meta-policy can wait until the burrowing exploration regions align with the assigned rewards and then select the policies trained on these rewards.
This usually ends up being more efficient than waiting for the agents to explore the whole map using independent rewards and can minimize the likelihood of detachment~\citep{ecoffet2019go}.

\end{document}